\documentclass[3p,10pt]{elsarticle}

\usepackage{amsmath,amssymb}
\usepackage{graphicx}
\usepackage{booktabs}
\usepackage{multirow}
\usepackage{threeparttable}
\usepackage{lineno}
\usepackage{url}
\usepackage{subcaption}
\usepackage{hyperref}

\journal{Biomedical Signal Processing and Control}

\begin{document}

\begin{frontmatter}

\title{Cross-Model Agreement as a Deployment-Time Reliability Signal for Automatic Polyp Segmentation}

\author[iitr]{Siddharth Gupta}
\author[iitr]{Jitin Singla\corref{cor1}}
\ead{jsingla@bt.iitr.ac.in}
\ead[url]{https://www.jitinsingla.in/} 
\cortext[cor1]{Corresponding author}
\affiliation[iitr]{organization={Department of Biosciences and Bioengineering, Indian Institute of Technology Roorkee},
            city={Roorkee},
            postcode={247667},
            state={Uttarakhand},
            country={India}}

\begin{abstract}
In real-time colonoscopy, ground-truth annotations are unavailable at inference, so polyp segmentation models can fail silently. We propose Referee-Based Quality Estimation (RBQE), a reference-free framework measuring agreement between a primary segmentation model and an independently trained referee on the same image. RBQE is evaluated on a standardized 1,223-image external benchmark drawn from four public datasets, using four referee configurations chosen to separate two design axes: referee independence and architectural diversity. Using a common Agreement Dice descriptor, a same-architecture referee differing from the primary model only in random initialization already yields a useful reliability signal (ROC-AUC $= 0.923$), showing that independent training alone is sufficient. Cross-architecture referees improve further: SegFormer-B0 achieves the strongest performance (ROC-AUC $= 0.960$), significantly outperforming the same-architecture control and UNet++, and exceeding a representative Test-Time Augmentation baseline by 0.055 ROC-AUC under an identical protocol, whereas a prompt-coupled MedSAM referee underperforms despite maximal architectural diversity. Because empty-mask agreement is trivially separable, we also report a restricted evaluation excluding such cases: ROC-AUC falls to 0.876 (SegFormer-B0, 1,046 images) and 0.783 (same-architecture control, 975 images), yet RBQE's margin over both baselines widens on this identical subset. RBQE additionally increases the mean Dice of retained predictions as low-agreement cases are progressively rejected, supporting selective prediction, and requires only one additional deterministic referee forward pass at inference. Our study therefore supports cross-model agreement as a practical, interpretable reliability framework for automated polyp segmentation.
\end{abstract}


\end{frontmatter}

\section{Introduction}
\label{sec:intro}

Colorectal cancer (CRC) is the third most commonly diagnosed cancer and the second leading cause of cancer-related death worldwide, with an estimated 1.93 million new cases and 904{,}000 deaths reported globally in 2022 \citep{Bray2024}. Timely detection and removal of precancerous polyps substantially improves patient survival, and colonoscopy remains the gold-standard technique for early polyp detection. However, colonoscopy is resource-intensive, requires specially trained endoscopists, and its diagnostic yield is highly dependent on operator experience --- a meta-analysis of 43 tandem-colonoscopy studies ($>$15{,}000 procedures) reported an adenoma miss rate of 26\% (95\% CI: 23--30\%) \citep{Zhao2019}.

Deep learning has substantially advanced automatic polyp segmentation, enabling more accurate lesion delineation and supporting computer-aided diagnosis (CAD) systems \citep{Jha2021}. Despite this progress, segmentation models continue to fail under challenging imaging conditions such as camera motion, poor illumination, specular artifacts, blur, very small polyps, and domain shift across imaging hardware. Because ground-truth annotations are unavailable at inference, such failures cannot be detected using standard metrics like the Dice Similarity Coefficient (DSC) or Intersection-over-Union (IoU), leaving clinicians and downstream decision-support systems with no reliable mechanism for flagging an unreliable segmentation before it informs diagnosis.

This gap has motivated no-reference Segmentation Quality Estimation (SQE), whose objective is to estimate segmentation reliability when ground-truth annotations are unavailable at inference. As detailed in Section~\ref{sec:related}, existing SQE approaches derive their quality signal either from a single model's own behavior (confidence scores, predictive entropy, Monte Carlo (MC) Dropout, Test-Time Augmentation(TTA)), from disagreement among ensembles of identically architected models, or from per-image reverse-classifier fitting against a reference atlas; these families respectively inherit the deployed model's calibration errors, a shared architectural inductive bias, and a substantial per-image retraining cost. To date, agreement between architecturally heterogeneous, independently trained segmentation models --- decoupled from any single model's calibration and requiring no per-image retraining --- has not been systematically investigated as a deployment-time reliability signal for polyp segmentation.

In the present work, we propose Referee-Based Quality Estimation (RBQE), a simple and practical framework that estimates segmentation reliability using agreement between independently trained segmentation models. Rather than relying on a single model's internal confidence, RBQE assesses the coherence between a primary segmentation model and an independently trained referee model applied to the same image. The underlying hypothesis is that when independently trained models converge to similar segmentation masks, the prediction is more likely to be reliable, whereas substantial disagreement indicates an increased likelihood of segmentation failure. We frame the choice of referee along two design axes: (i) \emph{independence} --- whether the referee's training and, crucially, its inference-time output are decoupled from the primary model --- and (ii) \emph{architectural diversity} --- whether the referee embodies a different inductive bias. The four referee configurations evaluated in this study (Section~\ref{sec:referees}) are chosen to separate these axes: a same-architecture, independently trained control isolates the contribution of independence; two architecturally distinct, independently trained referees add diversity; and a prompt-driven foundation model that is architecturally diverse but functionally coupled to the primary prediction isolates the role of output-level independence. RBQE is not proposed to predict the exact Dice coefficient, nor does it seek to replace supervised quality estimation models; instead, it provides a practical, deployment-time reliability estimate capable of flagging unreliable predictions when ground-truth labels are unavailable.

To evaluate this hypothesis, external validation was performed on a standardized benchmark of 1,223 colonoscopy images drawn from four independent, publicly available datasets, with all headline results additionally reported on restricted subsets that exclude trivially separable empty-mask cases. The proposed framework was compared against representative uncertainty-based (TTA) and morphology-based quality estimation baselines under a standardized evaluation protocol, using ROC-AUC, bootstrap confidence intervals, paired DeLong significance testing, and computational complexity analysis.

The principal contributions of this work are as follows:
\begin{enumerate}
\item \textbf{Isolating the role of referee independence.} Using a same-architecture control referee that differs from the primary model only in random initialization, we demonstrate that independent training alone, without architectural diversity, already provides a meaningful reliability signal (ROC-AUC $= 0.923$) --- to the best of our knowledge, a distinction that has not been explicitly investigated in the SQE literature.
\item \textbf{Cross-architecture diversity adds a statistically significant increment.} Agreement between the primary model and an independently trained, architecturally distinct referee (SegFormer-B0) provides the strongest deployment-time failure-detection signal (ROC-AUC $= 0.960$), significantly outperforming the same-architecture control and UNet++, and outperforming representative single-model uncertainty (TTA) and morphology-based baselines.
\item \textbf{Diversity without output-level independence fails.} A prompt-driven foundation-model referee (MedSAM, ROC-AUC $= 0.863$), whose inference prompt is derived from the primary model's own prediction, underperforms even the same-architecture control despite maximal architectural diversity --- identifying genuine output-level independence, rather than architectural difference alone, as the necessary ingredient of the agreement signal.
\item \textbf{A dual full/non-degenerate evaluation protocol.} Because empty-mask agreement is trivially separable, all headline comparisons are reported both on the full standardized benchmark and on restricted subsets excluding degenerate cases (ROC-AUC $0.960 \rightarrow 0.876$ for SegFormer-B0; $0.923 \rightarrow 0.783$ for the same-architecture control); RBQE's margin over the evaluated baselines widens, rather than narrows, on the restricted subset.
\item \textbf{Practical, label-free, single-pass deployment.} RBQE requires only one additional deterministic forward pass through a referee model at inference, with no per-image retraining, no quality-labelled training data, and no modification of the deployed primary model; five interpretable agreement descriptors are evaluated and ranked, and the continuous agreement score is shown to support selective prediction.
\end{enumerate}

\section{Related Work}
\label{sec:related}

In real-time colonoscopy, ground-truth annotations are unavailable during inference, making it impossible to directly assess whether an automatically generated segmentation is reliable. A substantial body of work has therefore investigated Segmentation Quality Estimation (SQE) without access to ground truth.

\subsection{Single-model quality signals}
The first family relies on the behavior of a single segmentation model. Robinson et al. \cite{Robinson2018} introduced a real-time CNN-based method for predicting segmentation quality directly from image and mask features for cardiac MRI quality control. Jungo and Reyes \cite{Jungo2019} systematically evaluated multiple uncertainty estimation methods for medical image segmentation and showed that existing approaches are reliable at the dataset level but often miscalibrated at the subject level. Eaton Rosen et al. \cite{EatonRosen2018} proposed a Bayesian deep learning framework converting voxel-wise uncertainty into calibrated volumetric confidence intervals. These approaches, together with MC Dropout \cite{Gal2016} and TTA \cite{Wang2019}, share a common limitation: the quality signal is derived entirely from one model's own behavior and therefore remains fundamentally dependent on that model's internal calibration; when a model is overconfident or encounters out-of-distribution data, these signals become unreliable precisely when accurate quality assessment matters most.

\subsection{Agreement-based estimation}
A second line of research instead estimates segmentation quality through agreement between multiple predictions. Deep ensembles \cite{Lakshminarayanan2017} train several instances of the same architecture from different random initializations and use their predictive disagreement as an uncertainty signal; random initialization alone can already yield substantial function-space diversity \cite{Fort2019}, yet ensemble members sharing an identical architectural inductive bias may still exhibit correlated rather than genuinely independent failure modes. Reverse Classification Accuracy (RCA) \cite{Valindria2017} instead fits a new classifier per test image using the model's own prediction as pseudo ground truth and scores it against a reference atlas, avoiding a second trained model but at the cost of per-image retraining and atlas dependence. RBQE is positioned within this agreement-based family but differs from both: it measures agreement between a fixed primary model and a single, independently trained referee requiring no retraining at inference time, and explicitly isolates whether the agreement signal depends on architectural diversity or on independent training alone --- a question not directly addressed by either deep ensembles or RCA.

\subsection{Selective prediction}
Beyond binary reliability estimation, selective classification with a reject option \cite{Chow1970,ElYaniv2010,Geifman2017} formalizes the risk--coverage trade-off between prediction coverage and error rate; RBQE's continuous agreement score is evaluated under this same framework in Section~\ref{sec:deployment}.

\section{The RBQE Framework}
\label{sec:method}

\subsection{Overview}
\label{sec:overview}

RBQE estimates reliability by measuring the agreement between two independently trained segmentation models applied to the same input image. Depending on the referee configuration, the models may share the same architecture but differ in initialization, or may additionally differ in architecture and training procedure. The underlying objective is to reduce the likelihood of systematically correlated failure modes while providing an independent prediction against which the primary model can be checked.

Let $I$ be the input colonoscopy image and $M_p$ be the primary segmentation model. Then
\begin{equation}
S_p = M_p(I),
\label{eq:primary}
\end{equation}
where $S_p$ denotes the binary segmentation mask produced by the primary model, deployed in clinical practice. An independently trained referee model, denoted $M_r$, processes the same image:
\begin{equation}
S_r = M_r(I).
\label{eq:referee}
\end{equation}

\begin{figure}[t]
\centering
\includegraphics[width=0.9\linewidth]{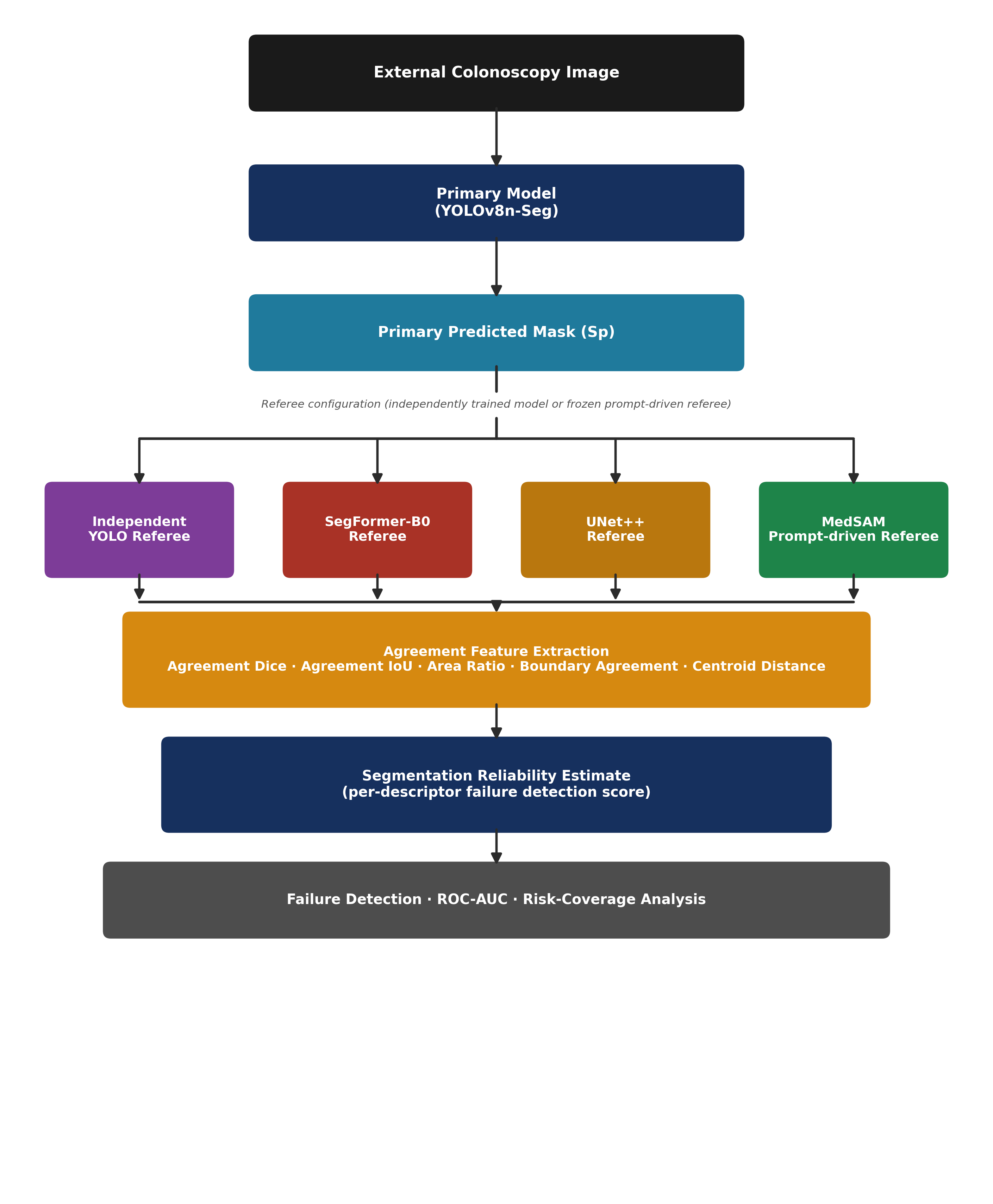}
\caption{Overview of the Referee-Based Quality Estimation (RBQE) framework. The primary model's prediction is compared against one of four referee configurations: an independently trained Independent YOLO Referee (same-architecture control), independently trained cross-architecture referees (SegFormer-B0 and UNet++), or a frozen prompt-driven MedSAM referee. Five agreement descriptors are then computed to obtain a deployment-time segmentation reliability estimate.}
\label{fig1:rbqe_framework}
\end{figure}

RBQE quantifies the consistency between $S_p$ and $S_r$ using a set of complementary agreement descriptors, described in Section~\ref{sec:features}. As illustrated in Fig.~\ref{fig1:rbqe_framework}, the framework consists of four steps: (i) generate a primary segmentation using the deployed model; (ii) obtain an independent prediction from a separately trained referee model; (iii) extract agreement features from the two masks; and (iv) estimate segmentation reliability from the resulting cross-model agreement.

RBQE operates directly on the output masks $S_p$ and $S_r$; it therefore requires no additional training, no architectural modification of the primary model, and no ground-truth annotations at inference. It is designed as a self-contained, post-hoc reliability module that can, in principle, be attached to any deployed segmentation pipeline producing a binary or probabilistic mask --- though this study validates it specifically with the architectures described in Sections~\ref{sec:primary}--\ref{sec:referees}. The complete evaluation protocol, including benchmark construction, the definition of segmentation failure, and the statistical testing procedures, is described in Section~\ref{sec:protocol}.

\subsection{Agreement descriptors}
\label{sec:features}

RBQE characterizes agreement between the primary and referee segmentation masks using descriptors that capture regional overlap, size consistency, boundary similarity, and spatial localization, summarized in Table~\ref{tab:features}.

\begin{table}[t]
\centering
\caption{Summary of RBQE agreement features.}
\label{tab:features}
\small
\resizebox{\textwidth}{!}{%
\begin{tabular}{llll}
\toprule
Feature & Captures & Range & Interpretation \\
\midrule
Agreement Dice & Regional overlap & $[0,1]$ & Higher $=$ stronger agreement \\
Agreement IoU & Regional overlap (stricter) & $[0,1]$ & Monotonic in Dice; comparability only \\
Area Ratio & Size consistency & $[0,1]$ & Higher $=$ similar predicted extent \\
Boundary Agreement & Contour consistency & $[0,1]$ & Higher $=$ aligned boundaries \\
Centroid Distance & Spatial localization & $\approx [0,1]$ & Lower $=$ closer agreement \\
\bottomrule
\end{tabular}}
\end{table}

\paragraph{Agreement Dice} Agreement Dice measures the overlap between the primary and referee segmentation masks,
\begin{equation}
D_{agr} = \frac{2\,|S_p \cap S_r|}{|S_p| + |S_r|},
\label{eq:agrdice}
\end{equation}
where $|S|$ denotes the number of foreground pixels in mask $S$. Larger values indicate stronger regional agreement.

\paragraph{Agreement IoU (remark)} Agreement IoU, $IoU_{agr} = |S_p \cap S_r| / |S_p \cup S_r|$, is reported for interpretability and comparability with prior segmentation literature. Note that $IoU_{agr}$ is a strictly monotonic function of $D_{agr}$ for any pair of masks,
\begin{equation}
IoU_{agr} = \frac{D_{agr}}{2 - D_{agr}},
\label{eq:iourelation}
\end{equation}
and therefore the two descriptors yield identical rank orderings, and consequently identical ROC-AUC values for failure detection and identical Spearman correlations with ground-truth segmentation quality (Section~\ref{sec:rq1}). Agreement IoU is retained alongside Agreement Dice for reader familiarity, rather than as an independent source of evidence; RBQE thus rests on four functionally distinct descriptors.

\paragraph{Area Ratio} To evaluate consistency in predicted lesion size, independent of spatial overlap, the Area Ratio is computed as
\begin{equation}
A_{ratio} = \frac{\min(|S_p|, |S_r|)}{\max(|S_p|, |S_r|)}.
\label{eq:arearatio}
\end{equation}
This feature is insensitive to small spatial shifts while capturing differences in foreground extent.

\paragraph{Boundary Agreement} Boundary Agreement evaluates contour consistency between the predicted masks by computing the F1-score between their extracted object boundaries, using a distance tolerance of 3 pixels to determine correspondence between boundary points, following the boundary F-measure formulation \cite{Perazzi2016}. Unlike region-based metrics, this descriptor emphasizes boundary alignment and captures fine contour discrepancies that may not substantially affect overlap-based measures.

\paragraph{Centroid Distance} Spatial consistency is evaluated using the Euclidean distance between the centroids of the predicted masks,
\begin{equation}
d_c = \sqrt{(x_p - x_r)^2 + (y_p - y_r)^2},
\label{eq:centroid}
\end{equation}
where $(x_p, y_p)$ and $(x_r, y_r)$ denote the centroids of the primary and referee segmentations, respectively. Centroid distances are normalized by the image diagonal, yielding a dimensionless quantity approximately within $[0,1]$ that is independent of image resolution and directly comparable across datasets. Smaller centroid distances indicate greater localization agreement.

\paragraph{Handling of degenerate cases} For images in which the primary or referee mask contains no foreground pixels, agreement features are assigned as follows: Agreement Dice $= 0$, Agreement IoU $= 0$, Boundary Agreement $= 0$, Area Ratio $= 0$ ($1.0$ if both masks are empty, reflecting identical zero extent), and Centroid Distance $= 1.0$ (worst case). Because such degenerate cases are trivially separable from substantive predictions, they motivate the dual full/non-degenerate evaluation protocol defined in Section~\ref{sec:benchmark}.

Together, these agreement descriptors characterize complementary aspects of segmentation consistency --- regional overlap, object size, boundary fidelity, and lesion localization --- with the exception of the Dice--IoU pair, which are mathematically related rather than independent. Rather than estimating the true Dice score, RBQE interprets the degree of agreement between independent segmentation models as a surrogate indicator of prediction reliability.

\subsection{Primary segmentation model}
\label{sec:primary}

YOLOv8n-Seg \citep{Jocher2023} serves as the primary segmentation model throughout this study; its predictions constitute the candidate segmentations whose reliability is subsequently estimated using RBQE. YOLOv8n-Seg was selected owing to its favorable balance between segmentation accuracy, computational efficiency, and real-time inference capability. The model simultaneously predicts object localization and pixel-wise segmentation masks within a unified architecture, making it suitable for deployment in resource-constrained clinical environments.

The primary segmentation model was trained exclusively on the Kvasir-SEG dataset (1,000 images) \citep{Jha2020} using the official Ultralytics implementation. The dataset was randomly partitioned into 700 training, 200 validation, and 100 held-out test images (70/20/10 split). Only the training and validation subsets were used during model development: the 700 training images were used for optimization and the 200 validation images were used exclusively for model selection. The 100-image held-out subset was not used for training or model selection and was not included in the external RBQE benchmark. The split was not stratified, as Kvasir-SEG is a single-class binary segmentation dataset. During training, standard data augmentation techniques such as random horizontal flipping, scaling, translation, and color-space augmentation were employed to improve generalization. Images were resized to the input resolution required by the network while preserving the corresponding binary segmentation masks. Unless otherwise specified, all training hyperparameters followed the recommended default configuration provided by the Ultralytics implementation, and the checkpoint achieving the highest validation segmentation performance was retained for all subsequent experiments.

During deployment, each external colonoscopy image was processed once by the primary segmentation model to generate the predicted binary segmentation mask. Importantly, RBQE does not modify or retrain the primary segmentation model. It functions as an independent, deployment-time quality estimation module that operates alongside existing segmentation systems without altering their training or inference procedures. This property is what allows RBQE to be evaluated against multiple referee configurations (Section~\ref{sec:referees}) without any corresponding change to the primary model itself.

\subsection{Referee configurations: independence and architectural diversity}
\label{sec:referees}

To evaluate whether the proposed agreement signal depends on architectural diversity, model independence, or both, four referee configurations were investigated, arranged along the two design axes introduced in Section~\ref{sec:intro}: one same-architecture, independently trained control; two cross-architecture, independently trained referees; and one cross-architecture, prompt-coupled control. The Independent YOLO Referee, SegFormer-B0, and UNet++ were independently trained on Kvasir-SEG, whereas MedSAM was used as a frozen pretrained foundation model. Because MedSAM requires a prompt derived from the primary prediction, it is treated as a prompt-dependent architectural control rather than as a fully independent referee.

\paragraph{Independent YOLO Referee (independent / same architecture)} A second YOLOv8n-Seg model was trained using an architecture and training protocol identical to the primary segmentation model, but with an independent random weight initialization and optimization trajectory. This model shares no weights, training checkpoints, or optimization state with the primary model and was never used to produce the primary segmentation. Its role is exclusively that of a referee at deployment, providing a same-architecture control condition that isolates the contribution of model independence from that of architectural diversity.

\paragraph{SegFormer-B0 (independent / cross-architecture)} SegFormer-B0 \citep{Xie2021} represents a transformer-based semantic segmentation architecture employing a hierarchical Mix Transformer encoder with a lightweight decoder. Owing to its strong segmentation capability and architectural independence from the primary model, SegFormer-B0 serves as the principal cross-architecture referee throughout this study.

\paragraph{UNet++ (independent / cross-architecture)} UNet++ \citep{Zhou2018} represents a convolutional encoder--decoder architecture with nested dense skip connections. This model is included to investigate whether RBQE remains effective when the referee belongs to a fundamentally different, CNN-based segmentation family, distinct from both the primary model and SegFormer-B0.

\paragraph{MedSAM (prompt-coupled / cross-architecture)} MedSAM \citep{Ma2024} represents a prompt-driven medical foundation model derived from the Segment Anything framework. Since MedSAM requires external prompts during inference --- here, a bounding-box prompt derived from the primary model's predicted mask --- its output is not fully independent of the primary model, despite maximal architectural diversity. It is therefore employed to isolate the influence of output-level independence in prompt-based segmentation systems, rather than as a principal cross-architecture referee.

Table~\ref{tab:refereetraining} summarizes the training configuration for each referee. The Independent YOLO Referee, SegFormer-B0, and UNet++ were each trained from a pretrained initialization on the Kvasir-SEG dataset, following the same training/validation protocol described for the primary model (Section~\ref{sec:primary}). SegFormer-B0 and UNet++ were each trained from ImageNet-pretrained encoders using their respective official implementations, with hyperparameters selected according to common practice for each architecture. MedSAM was used as a frozen, officially released pretrained checkpoint, applied directly at inference without any additional training or fine-tuning on Kvasir-SEG or the external evaluation datasets. All four referee models were trained or evaluated exclusively using the Kvasir-SEG dataset (for the three trainable referees) or the officially released pretrained checkpoint (for MedSAM), and none were fine-tuned or adapted on any of the four external evaluation datasets (Section~\ref{sec:datasets}), consistent with the cross-dataset generalization protocol applied to the primary model.

\begin{table}[t]
\centering
\caption{Training configuration for each referee model. MedSAM was used as a frozen pretrained checkpoint and was not trained or fine-tuned on any dataset in this study.}
\label{tab:refereetraining}
\footnotesize
\setlength{\tabcolsep}{3.5pt}
\resizebox{\textwidth}{!}{%
\begin{tabular}{lllllllll}
\toprule
Referee & Architecture & Pretraining & Epochs & Optimizer & LR & Batch & Resolution & Framework \\
\midrule
Independent YOLO & YOLOv8n-Seg & YOLOv8n-Seg pretrained; & 100 & SGD (default) & default & 16 & $640{\times}640$ & Ultralytics YOLOv8 \\
Referee & & independent run & & & & & &  \\
SegFormer-B0 & SegFormer-B0 & ImageNet & 100 & AdamW & $6{\times}10^{-5}$ & 8 & $512{\times}512$ & MMSegmentation \\
UNet++ & UNet++ (ResNet34) & ImageNet & 100 & Adam & $1{\times}10^{-4}$ & 8 & $512{\times}512$ & segmentation\_models.pytorch \\
MedSAM & SAM-based & Official pretrained & --- & --- & --- & --- & $1024{\times}1024$ & Official MedSAM \\
 & foundation model & checkpoint & & & & & & implementation \\
\bottomrule
\end{tabular}}
\end{table}

The objective of evaluating these four referee configurations is not to compare segmentation architectures for their own sake, but to determine whether agreement-based quality estimation depends on architectural diversity between the primary and referee models, on independent model training alone, or on a combination of both.

\section{Experimental Protocol}
\label{sec:protocol}

This section describes the datasets, benchmark construction, evaluation metrics, baselines, and implementation details used to assess the proposed framework. Unless otherwise specified, all experiments were performed using identical preprocessing, evaluation criteria, and deployment settings to ensure fair comparison across referee architectures and competing quality estimation methods.

\subsection{Datasets}
\label{sec:datasets}

The proposed framework was developed using the Kvasir-SEG dataset \citep{Jha2020} and evaluated exclusively on four independent public colonoscopy datasets. Using external datasets for evaluation enables assessment of the framework under realistic domain shifts arising from differences in imaging equipment, acquisition protocols, patient populations, bowel preparation quality, illumination conditions, and polyp characteristics. The Kvasir-SEG dataset was used solely for training the primary segmentation model and the referee segmentation models; no images from Kvasir-SEG were included in the final evaluation benchmark.

The external benchmark comprises the following publicly available datasets, summarized in Table~\ref{tab:datasets}:
\begin{itemize}
\item \textbf{CVC-ClinicDB} \citep{Bernal2015} contains 612 colonoscopy images with manually annotated polyp segmentation masks acquired during routine clinical examinations, exhibiting relatively high image quality and moderate variation in polyp appearance.
\item \textbf{CVC-ColonDB} \citep{Bernal2012} consists of 380 colonoscopy images containing numerous challenging cases characterized by small polyps, irregular boundaries, specular highlights, and complex mucosal textures.
\item \textbf{ETIS-Larib PolypDB} \citep{Silva2014} contains 196 particularly difficult colonoscopy images with substantial variability in polyp size, shape, illumination, and image quality, widely regarded as one of the most challenging benchmarks for polyp segmentation.
\item \textbf{CVC-300} \citep{Vazquez2017} contains 60 annotated colonoscopy images, commonly used to evaluate model generalization under limited-data conditions.
\end{itemize}

Combining these four datasets results in a diverse external benchmark containing a broad spectrum of lesion sizes, imaging conditions, anatomical variations, and segmentation difficulty levels. Since none of these datasets were used during model training, they provide an unbiased assessment of deployment-time segmentation reliability under realistic clinical conditions.

\begin{table}[t]
\centering
\caption{Summary of the datasets used for training and external evaluation.}
\label{tab:datasets}
\small
\begin{tabular}{lrll}
\toprule
Dataset & Images & Role & Used for \\
\midrule
Kvasir-SEG & 1,000 & Training & Primary and referee model development \\
CVC-ClinicDB & 612 & Testing & External benchmark \\
CVC-ColonDB & 380 & Testing & External benchmark \\
ETIS-Larib PolypDB & 196 & Testing & External benchmark \\
CVC-300 & 60 & Testing & External benchmark \\
\bottomrule
\end{tabular}
\end{table}

\subsection{Benchmark construction and evaluation subsets}
\label{sec:benchmark}

For standalone segmentation evaluation, all 1,248 external images (Section~\ref{sec:datasets}) were retained to provide an unbiased assessment of the primary segmentation model. RBQE, however, estimates segmentation reliability by analyzing the agreement between the primary segmentation model and an independently trained referee model, and its evaluation therefore uses two nested definitions, both fixed \emph{a priori} and applied uniformly.

\paragraph{Full standardized benchmark ($N = 1{,}223$)} During benchmark construction, 25 images were identified in which both the primary model and the SegFormer-B0 referee — the first referee evaluated — produced empty segmentation masks. Since these cases contain no segmented foreground in either mask, they represent degenerate cases that provide no meaningful information regarding the ability of RBQE to distinguish reliable from unreliable segmentations; including them would artificially inflate agreement-based quality estimation metrics while contributing no evidence regarding segmentation correctness. These 25 images were therefore excluded, resulting in a standardized benchmark of 1,223 images. This identical, fixed benchmark was subsequently used for all four referee configurations, rather than computing referee-specific exclusion sets, ensuring that all referee models were evaluated on exactly the same images and that ROC-AUC, bootstrap confidence intervals, DeLong tests, and all other evaluation metrics remain directly comparable across referees without confounding differences in sample selection. Unless explicitly stated otherwise, all reported RBQE performance metrics correspond to this standardized 1,223-image benchmark.

\paragraph{Non-degenerate (restricted) subsets} Because RBQE's agreement descriptors can be trivially informative when a mask is empty (Section~\ref{sec:features}), headline results are additionally reported on restricted subsets excluding degenerate cases. Empty-mask status is referee-specific, so the restricted subsets differ in composition and size across referees. For the SegFormer-B0 configuration, all images in which either the primary or the referee mask contains no more than a negligible foreground region were excluded; this removed 177 images --- all corresponding to a primary-empty prediction with a non-empty SegFormer-B0 referee mask (177 primary-empty, 0 referee-empty) --- yielding a restricted subset of 1,046 images in which both masks contain substantive predicted foreground. For the Independent YOLO Referee, the corresponding restricted subset is its ``Both Non-Empty'' stratum of 975 images (Section~\ref{sec:rq3}). Importantly, the 130 ``Both Empty'' cases in the Independent YOLO stratification are specific to that referee and are not the 25 SegFormer-B0 double-empty cases excluded during benchmark construction; they remained in the standardized benchmark because benchmark membership was determined once, using the SegFormer-B0 exclusion criterion, before evaluating the other referee configurations.

\subsection{Failure definition and evaluation metrics}
\label{sec:metrics}

The effectiveness of RBQE was evaluated as a binary segmentation failure detection problem. Given a predicted segmentation mask and its corresponding ground-truth annotation, segmentation quality was quantified using the Dice Similarity Coefficient (DSC),
\begin{equation}
DSC = \frac{2\,|S_p \cap G|}{|S_p| + |G|},
\label{eq:dsc}
\end{equation}
where $S_p$ denotes the predicted segmentation mask and $G$ the corresponding ground-truth mask. This is the same overlap formulation as Agreement Dice (Eq.~\ref{eq:agrdice}), applied here against ground truth rather than between the primary and referee predictions. Following prior segmentation quality estimation literature \citep{Valindria2017}, a prediction was classified as a segmentation failure when
\begin{equation}
DSC < 0.50.
\label{eq:failure}
\end{equation}
This threshold serves as the primary operating point throughout the paper, while additional threshold sensitivity analyses (Section~\ref{sec:rq3}) assess robustness under alternative failure definitions.

Since RBQE produces a continuous reliability score rather than a binary prediction, its discriminative ability was primarily assessed using the Area Under the Receiver Operating Characteristic Curve (ROC-AUC), selected because it is threshold-independent and measures the ability of the quality estimation method to distinguish reliable from unreliable segmentations across all possible operating thresholds. For threshold-dependent evaluation, binary decisions were obtained using the Youden-optimal operating threshold \citep{Youden1950} determined from the corresponding ROC curve. Performance was subsequently quantified using
\begin{align}
\text{Accuracy} &= \frac{TP + TN}{TP + TN + FP + FN}, \label{eq:acc}\\
\text{Precision} &= \frac{TP}{TP + FP}, \qquad
\text{Recall} = \frac{TP}{TP + FN}, \label{eq:precrec}\\
F_1 &= 2 \times \frac{\text{Precision} \times \text{Recall}}{\text{Precision} + \text{Recall}}, \label{eq:f1}
\end{align}
where $TP$, $TN$, $FP$, and $FN$ denote the numbers of true positives, true negatives, false positives, and false negatives, respectively.

To quantify statistical uncertainty, 95\% confidence intervals for ROC-AUC were estimated using 1,000 bootstrap resamples drawn with replacement at the image level from the standardized 1,223-image benchmark. Pairwise comparisons between correlated ROC curves were performed using DeLong's test \citep{DeLong1988} to determine whether observed differences between competing quality estimation methods were statistically significant. A significance level of $p < 0.05$ was adopted throughout the study.

\subsection{Baselines}
\label{sec:baselines}

\paragraph{Test-Time Augmentation (TTA)} For comparison with single-model uncertainty estimation, a TTA baseline was evaluated using the same primary YOLOv8n-Seg model and the standardized external benchmark. Nine inference conditions were considered: horizontal flipping, increased brightness, decreased brightness, increased contrast, decreased contrast, gamma correction ($\gamma = 1.1$), CLAHE, Gaussian blur, and the unmodified input image. Each transformed image was processed by the primary segmentation model and the resulting predictions were converted to binary masks. The resulting prediction set was used to derive uncertainty descriptors, including mean predictive entropy, maximum predictive entropy, zero-prediction count, and pixel-level disagreement ratio. The zero-prediction count provided the strongest failure-detection performance among the evaluated TTA descriptors and was therefore used as the TTA reliability signal in the baseline comparison.

\paragraph{Morphology-based SQE} A morphology-based baseline derives a quality signal from geometric properties of the predicted mask alone; among the evaluated geometric descriptors, Circularity achieved the strongest failure-detection performance and is used as the morphology-based baseline.

Both baselines were evaluated using the same failure definition (ground-truth $DSC < 0.50$), standardized benchmark, ROC-AUC analysis, and statistical evaluation protocol used for RBQE, and were additionally recomputed on the identical 1,046-image restricted subset used for the SegFormer-B0 configuration (Section~\ref{sec:rq4}).

\subsection{Implementation details}
\label{sec:implementation}

All models were implemented using the PyTorch deep learning framework. The primary segmentation model (YOLOv8n-Seg) was trained using the official Ultralytics implementation, while SegFormer-B0, UNet++, MedSAM, and the Independent YOLO Referee were implemented using their respective official or publicly available repositories. Unless otherwise specified, the original network architectures and recommended training configurations were adopted for all models to ensure fair comparison (Table~\ref{tab:refereetraining}).

All input colonoscopy images were resized to the resolution required by the corresponding segmentation model before inference. Predicted probability maps were resampled to the original input image's resolution prior to thresholding --- using bilinear interpolation for the continuous logits (SegFormer-B0, UNet++) and nearest-neighbor resizing for already-binarized outputs (YOLOv8n-Seg) --- before being converted to binary segmentation masks using the default inference settings provided by the respective implementations, and these binary masks served as the direct inputs to the proposed RBQE framework. This ensures that all agreement features (Section~\ref{sec:features}) were computed on a common, per-image pixel grid regardless of each model's native inference resolution (Table~\ref{tab:refereetraining}), rather than across mismatched grids (e.g., 640$\times$640 for the primary model and Independent YOLO Referee, 512$\times$512 for SegFormer-B0 and UNet++, 1024$\times$1024 for MedSAM). Agreement features were computed directly from the binary masks generated by the primary model and each referee model. For each experiment, the continuous agreement score produced by RBQE was interpreted as a segmentation reliability score, with larger values indicating greater confidence in the predicted segmentation.

To ensure reproducibility, the same external benchmark, ground-truth failure definition, agreement-feature extraction procedures, and evaluation metrics were used across all referee architectures and competing quality-estimation methods. All experiments were performed on the workstation configuration summarized in Table~\ref{tab:implementation}, with model training and inference accelerated using CUDA.

\begin{table}[t]
\centering
\caption{Implementation details.}
\label{tab:implementation}
\small
\begin{tabular}{lp{9.2cm}}
\toprule
Framework & PyTorch \\
Primary model & YOLOv8n-Seg \\
Referee models & SegFormer-B0, UNet++, MedSAM, Independent YOLO Referee \\
Training dataset & Kvasir-SEG \\
Evaluation datasets & CVC-ClinicDB, CVC-ColonDB, ETIS-Larib PolypDB, CVC-300 \\
Operating system & Windows 11 \\
GPU & NVIDIA RTX 4060 (8 GB) \\
CPU & AMD Ryzen 7 7700X \\
RAM & 32 GB \\
Statistical analysis & Bootstrap (1,000 resamples), DeLong test \\
\bottomrule
\end{tabular}
\end{table}

\section{Results}
\label{sec:results}

Results are organized around four research questions. \textbf{RQ1:} Does cross-model agreement detect segmentation failure, and which agreement descriptors carry the signal (Section~\ref{sec:rq1})? \textbf{RQ2:} Does the signal require architectural diversity, or does independent training alone suffice --- and what happens when diversity is present but output-level independence is not (Section~\ref{sec:rq2})? \textbf{RQ3:} How robust is the signal to the definition of segmentation failure and to degenerate empty-mask cases (Section~\ref{sec:rq3})? \textbf{RQ4:} How does RBQE compare with representative existing no-reference SQE paradigms (Section~\ref{sec:rq4})? Sections~\ref{sec:deployment} and \ref{sec:qualitative} then examine deployment characteristics --- selective prediction and computational cost --- and qualitative behavior. Section~\ref{sec:primaryperf} first characterizes the primary model itself.

\subsection{External segmentation performance of the primary model}
\label{sec:primaryperf}

Since RBQE estimates the reliability of predicted segmentations rather than improving segmentation accuracy itself, it must be evaluated under realistic conditions where both successful and failed predictions naturally occur. We therefore first characterize the segmentation performance of the deployed primary model on previously unseen data. The primary YOLOv8n-Seg model was trained exclusively on the Kvasir-SEG dataset and evaluated without any fine-tuning on the four independent external datasets, which exhibit considerable variation in imaging conditions, lesion appearance, polyp size, illumination, and acquisition protocols. Table~\ref{tab:primaryperf} summarizes the segmentation performance on each external dataset.

\begin{table}[t]
\centering
\caption{External segmentation performance of YOLOv8n-Seg on unseen datasets.}
\label{tab:primaryperf}
\small
\begin{tabular}{lcccc}
\toprule
Dataset & Dice & IoU & Precision & Recall \\
\midrule
CVC-ClinicDB & 0.766 & 0.699 & 0.773 & 0.802 \\
CVC-ColonDB & 0.675 & 0.607 & 0.673 & 0.698 \\
ETIS-Larib PolypDB & 0.609 & 0.557 & 0.587 & 0.651 \\
CVC-300 & 0.828 & 0.749 & 0.769 & 0.906 \\
\midrule
Weighted mean (1,248 images) & 0.717 & 0.651 & 0.713 & 0.752 \\
\bottomrule
\end{tabular}
\end{table}

As expected, segmentation performance varies across the four datasets owing to their differing levels of difficulty. Images containing small polyps, irregular object boundaries, poor illumination, specular reflections, motion blur, and complex mucosal textures are generally more challenging than images containing well-defined lesions acquired under favorable imaging conditions. Consequently, the external benchmark contains a broad spectrum of segmentation quality, ranging from highly accurate predictions to complete segmentation failures. This diversity is essential for evaluating deployment-time quality estimation: a benchmark with uniformly accurate predictions would make quality estimation a trivial task, whereas the mix of successful and failed segmentations here provides a meaningful test bed for cross-model agreement.

\subsection{RQ1: Agreement as a failure-detection signal, and descriptor behavior}
\label{sec:rq1}

RBQE was first evaluated using SegFormer-B0 as the referee model, selected for its strong segmentation performance, architectural independence from the YOLOv8n-Seg primary model, and favorable computational efficiency. For every image in the standardized 1,223-image benchmark, both the primary segmentation model and the independently trained SegFormer-B0 referee generated binary segmentation masks. The agreement descriptors described in Section~\ref{sec:features} were computed from the two predicted masks, and failure-detection performance was assessed for each descriptor individually according to the protocol in Section~\ref{sec:metrics}, alongside the Spearman rank correlation of each descriptor with ground-truth segmentation quality (DSC). Table~\ref{tab:descriptors} summarizes both analyses.

\begin{table}[t]
\centering
\resizebox{\textwidth}{!}{%
\begin{threeparttable}
\caption{Failure-detection performance and correlation with ground-truth quality for individual agreement descriptors, using SegFormer-B0 as the referee on the full standardized 1,223-image benchmark. Threshold-dependent metrics use the Youden-optimal operating threshold. The best-performing descriptor is highlighted in bold.}
\label{tab:descriptors}
\small
\begin{tabular}{lcccccc}
\toprule
Agreement signal & ROC-AUC & Accuracy & Precision & Recall & F1 & Spearman $\rho$ \\
\midrule
\textbf{Agreement Dice} & \textbf{0.960} & \textbf{0.926} & \textbf{0.762} & \textbf{0.894} & \textbf{0.823} & \textbf{0.734} \\
Agreement IoU\tnote{a} & 0.960 & 0.926 & 0.762 & 0.894 & 0.823 & 0.734 \\
Centroid Distance & 0.951 & 0.925 & 0.744 & 0.881 & 0.807 & $-0.620$ \\
Boundary Agreement & 0.949 & 0.923 & 0.738 & 0.873 & 0.800 & 0.622 \\
Area Ratio & 0.463 & 0.804 & 0.493 & 0.356 & 0.413 & 0.131 \\
\bottomrule
\end{tabular}
\begin{tablenotes}\footnotesize
\item[a] Identical to Agreement Dice by construction (Eq.~\ref{eq:iourelation}); reported for comparability only.
\end{tablenotes}
\end{threeparttable}}
\end{table}

As expected from Eq.~\ref{eq:iourelation}, Agreement Dice and Agreement IoU produced identical rankings and therefore identical ROC-AUC and Spearman correlation values. Centroid Distance and Boundary Agreement each provide strong failure-detection signals close to the overlap-based descriptors. Area Ratio, by contrast, achieves a ROC-AUC of only 0.463 --- near chance level --- indicating that predicted lesion size alone, independent of spatial overlap or location, is a weak and unreliable indicator of segmentation failure at the primary $DSC < 0.50$ operating point; given this near-chance performance, Area Ratio is not recommended as a standalone reliability descriptor in deployment settings.

The Spearman analysis is consistent with this ranking: Agreement Dice and Agreement IoU achieve the strongest positive correlation with ground-truth DSC ($\rho = 0.734$), followed by Boundary Agreement ($\rho = 0.622$). Centroid Distance exhibits a strong negative correlation ($\rho = -0.620$), consistent with its interpretation as an error measure rather than an agreement measure: lower centroid distance corresponds to higher segmentation quality. Area Ratio shows only weak positive correlation ($\rho = 0.131$), consistent with its near-chance ROC-AUC. A segmentation can reproduce a lesion's approximate size while still being incorrectly localized or poorly delineated at the boundary, so size consistency alone says little about where and how the lesion is delineated; overlap- and boundary-based measures therefore serve as the primary reliability signal, while size- and location-based descriptors offer weaker but complementary information in specific failure regimes. RBQE's descriptors are thus not interchangeable.

\begin{figure}[t]
\centering
\begin{subfigure}[b]{0.48\linewidth}
    \includegraphics[width=\linewidth]{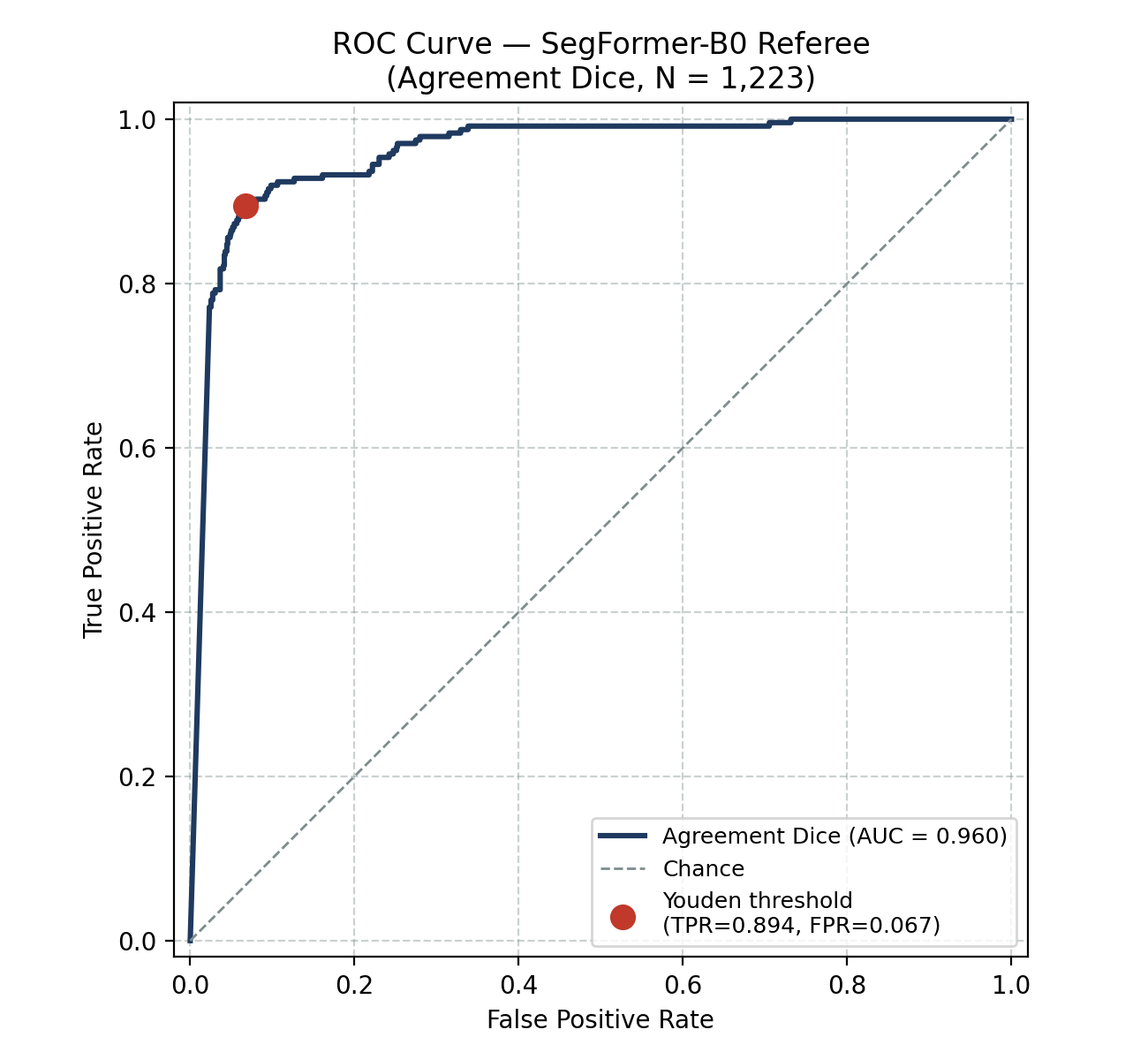}
    \caption{}
    \label{fig:roc-a}
\end{subfigure}\hfill
\begin{subfigure}[b]{0.48\linewidth}
    \includegraphics[width=\linewidth]{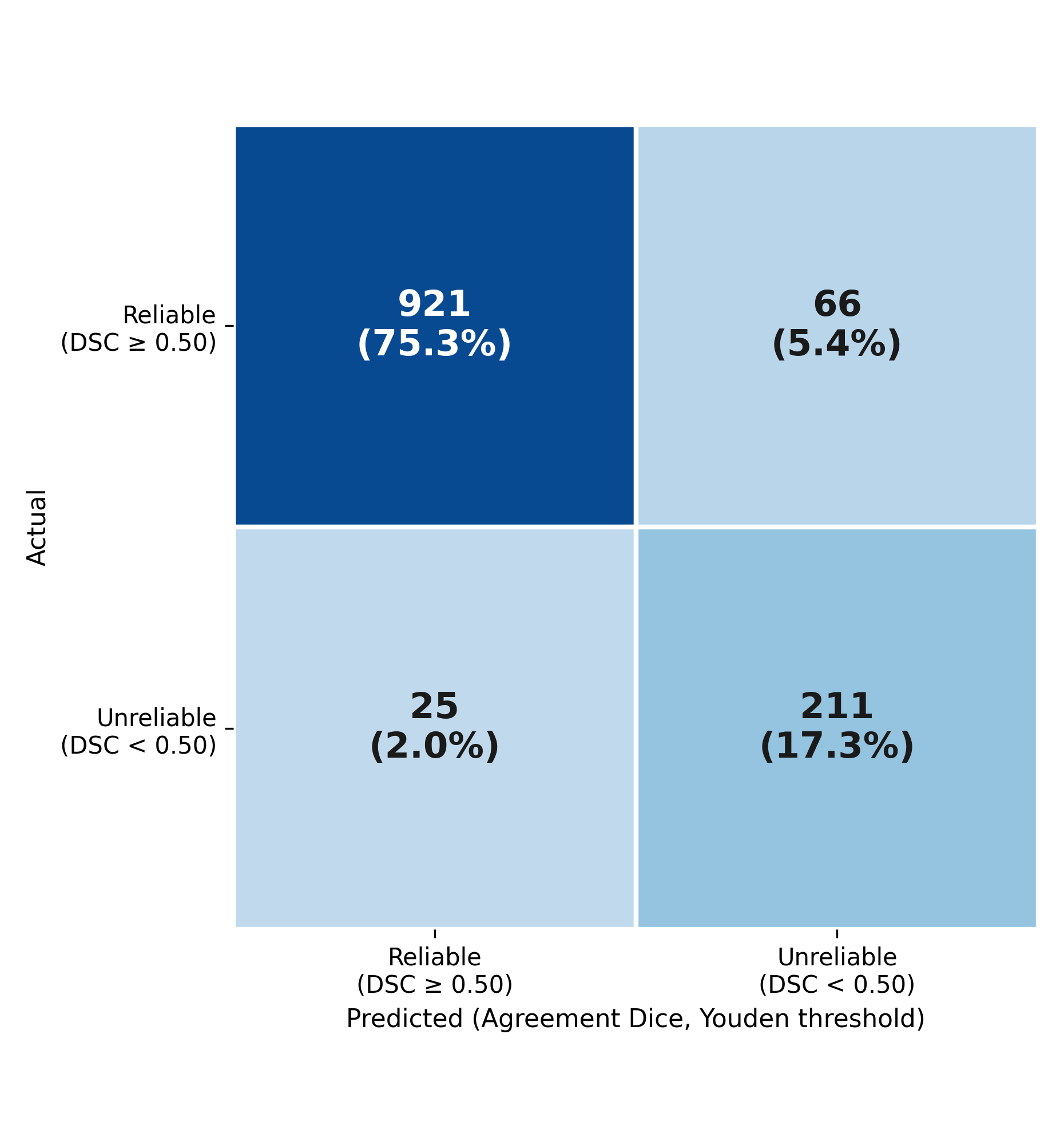}
    \caption{}
    \label{fig:roc-b}
\end{subfigure}
\caption{(a) Receiver Operating Characteristic (ROC) curve for Agreement Dice (SegFormer-B0 referee, $N = 1{,}223$), AUC $= 0.960$. The marked point indicates the Youden-optimal operating threshold used for the threshold-dependent metrics in Table~\ref{tab:descriptors}. (b) Confusion matrix for SegFormer-B0 referee failure detection using Agreement Dice at the Youden-optimal operating threshold ($N = 1{,}223$).}
\label{fig:roc}
\end{figure}

\begin{figure}[t]
\centering
\includegraphics[width=0.62\linewidth]{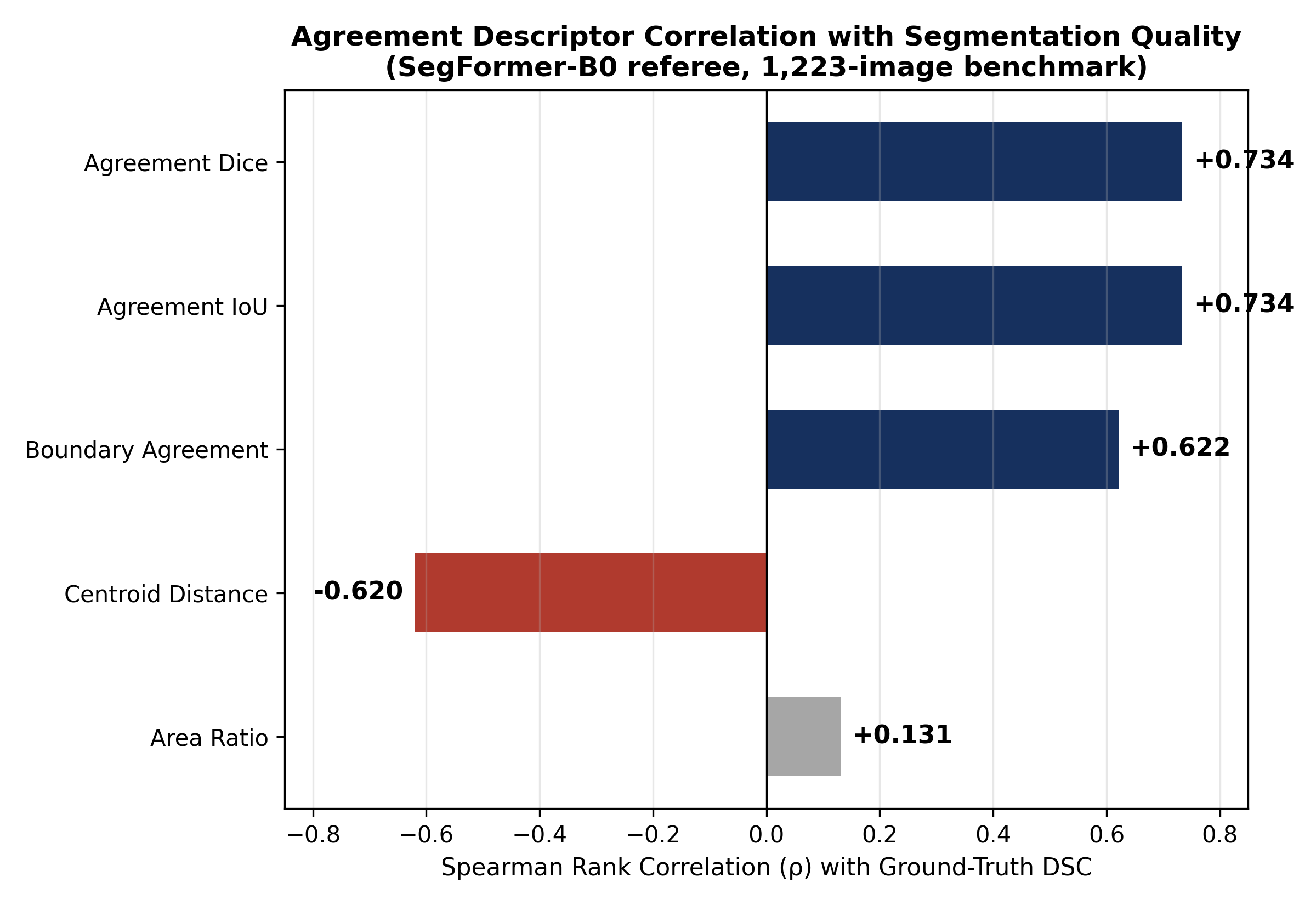}
\caption{Spearman rank correlation between each RBQE agreement descriptor and ground-
truth DSC (SegFormer-B0 referee, 1,223-image benchmark), corresponding to Table \ref{tab:descriptors}.}
\label{fig:scatter}
\end{figure}

Based on this comparison, Agreement Dice is adopted as the primary descriptor for all subsequent analyses. Fig.~\ref{fig:roc}a presents its ROC curve, and Fig.~\ref{fig:roc}b the confusion matrix at the Youden-optimal operating threshold, illustrating the distribution of correctly identified successful segmentations, correctly identified failures, false alarms, and missed failures. Fig.~\ref{fig:scatter} shows the association between the Agreement Dice value and the corresponding ground-truth Dice score for all images in the standardized benchmark: images with higher Agreement Dice values generally exhibit more accurate segmentations, whereas low agreement values are predominantly associated with segmentation failures. Although some overlap exists between successful and unsuccessful cases, reflecting the inherent variability of medical image segmentation, the overall trend is consistent with the strong discriminative performance in Table~\ref{tab:descriptors}. The corresponding non-degenerate (restricted-subset) performance of Agreement Dice is reported alongside the referee comparison in Section~\ref{sec:rq2}.

Overall, cross-model consistency provides an informative surrogate indicator of segmentation reliability: strong agreement between the independently trained primary and referee models generally corresponds to successful segmentation, while substantial disagreement is associated with failure. Unlike methods that derive confidence from repeated evaluation of a single model, RBQE estimates reliability from consistency between independent predictions.

\subsection{RQ2: Independence versus architectural diversity}
\label{sec:rq2}

An open question is whether the observed agreement principle requires architectural diversity between the primary and referee models, or whether independent model training alone --- without any architectural difference --- is sufficient to produce a useful reliability signal. RBQE was therefore evaluated using the four referee configurations of Section~\ref{sec:referees}. For each configuration, the complete RBQE pipeline remained unchanged; only the referee segmentation model was replaced, while identical agreement features, failure definitions, evaluation metrics, and experimental settings were maintained throughout. Consequently, any observed performance differences can be attributed solely to the characteristics of the referee model rather than to modifications of the quality estimation framework.

Table~\ref{tab:referees} presents the head-to-head comparison, using a fixed Agreement Dice descriptor applied uniformly across all referees rather than selecting the best-performing descriptor independently for each (the corresponding best-of-five descriptor results are reported later in this section as a sensitivity analysis; Table~\ref{tab:bestdescriptor}). The table consolidates, for each referee: full-benchmark ROC-AUC with bootstrap 95\% confidence intervals (1,000 resamples), the restricted-subset ROC-AUC on the referee's non-degenerate subset (Section~\ref{sec:benchmark}), the paired DeLong comparison against SegFormer-B0 using matched Agreement Dice scores, and threshold-dependent metrics at the Youden-optimal operating point.

\begin{table}[t]
\centering
\resizebox{\textwidth}{!}{%
\begin{threeparttable}
\caption{Primary head-to-head referee comparison using a fixed Agreement Dice descriptor, applied uniformly across all four referees. Full-benchmark results on the standardized 1,223-image benchmark, with bootstrap 95\% confidence intervals (1,000 resamples); restricted results on each referee's non-degenerate subset (Section~\ref{sec:benchmark}); DeLong $p$-values from paired comparisons against SegFormer-B0 using matched Agreement Dice scores. Threshold-dependent metrics (Acc., Prec., Rec., F1) at the Youden-optimal threshold on the full benchmark.}
\label{tab:referees}
\footnotesize
\setlength{\tabcolsep}{3pt}
\begin{tabular}{llccccccc}
\toprule
Referee & Category & \begin{tabular}[c]{@{}c@{}}ROC-AUC\\ full {[}95\% CI{]}\end{tabular} & \begin{tabular}[c]{@{}c@{}}ROC-AUC\\ restricted ($N$)\end{tabular} & \begin{tabular}[c]{@{}c@{}}DeLong $p$\\ vs.\ SegFormer\end{tabular} & Acc. & Prec. & Rec. & F1 \\
\midrule
SegFormer-B0 & Cross-arch. & 0.960 [0.947, 0.972] & 0.876 (1,046) & ref. & 0.926 & 0.762 & 0.894 & 0.823 \\
UNet++ & Cross-arch. & 0.938 [0.925, 0.951] & --- & $4.29 \times 10^{-3}$ & 0.863 & 0.590 & 0.941 & 0.725 \\
Independent YOLO & Same-arch.\ control & 0.923 [0.902, 0.942] & 0.783 (975) & $6.32 \times 10^{-5}$ & 0.908 & 0.710 & 0.881 & 0.786 \\
MedSAM\tnote{a} & Cross-arch.\ (prompt-coupled) & 0.863 [0.826, 0.902] & --- & n/a\tnote{b} & 0.952 & 1.000 & 0.750 & 0.857 \\
\bottomrule
\end{tabular}
\begin{tablenotes}\footnotesize
\item[a] MedSAM's Agreement Dice and Centroid Distance thresholds (${\approx}0$ and $1.0$, respectively) happen to flag the identical 177 images as unreliable ($TP{=}177$, $FN{=}59$, $FP{=}0$, $TN{=}987$), despite differing ROC-AUCs (0.863 vs.\ 0.873)---because a badly failed prompt-derived mask tends to diverge in overlap and centroid together. Area Ratio and Boundary Agreement give distinct confusion matrices under the same pipeline, ruling out a computation artifact.
\item[b] MedSAM was excluded from the paired DeLong comparison: its prompt-dependent inference precludes an independent paired signal.
\end{tablenotes}
\end{threeparttable}}
\end{table}

\paragraph{Independence alone suffices; diversity adds a significant increment} The Independent YOLO Referee (same-architecture control, ROC-AUC $= 0.923$) is outperformed by both SegFormer-B0 ($0.960$) and UNet++ ($0.938$), while exceeding MedSAM ($0.863$). The DeLong analysis confirms that SegFormer-B0 significantly outperforms both the Independent YOLO Referee ($p < 0.001$) and UNet++ ($p = 4.29 \times 10^{-3}$), establishing that the improvement associated with cross-architecture diversity is statistically significant rather than arising from random sampling variation. UNet++ also achieves a higher ROC-AUC than the same-architecture control, although a direct DeLong comparison between UNet++ and the Independent YOLO Referee was not performed. These results support the interpretation that independent training alone provides a useful reliability signal, while architectural diversity can provide additional discriminative strength when the referee remains genuinely independent of the primary model.

\paragraph{Diversity without output-level independence fails} MedSAM is a partial exception to this trend: despite being the most architecturally distinct of the four referees, it achieved the lowest ROC-AUC ($0.863$), underperforming even the same-architecture YOLO control. As established in Section~\ref{sec:referees}, this is attributable to MedSAM's prompt-driven inference paradigm: because its segmentation prompt is derived from the primary model's own predicted mask, its output is not fully independent of the primary model, despite the substantial architectural difference between a promptable foundation model and a convolutional detector. This finding refines the central hypothesis of this work: architectural diversity is beneficial only insofar as it is accompanied by genuine output-level independence; a referee that is architecturally distinct but functionally coupled to the primary model's predictions provides a weaker signal than a same-architecture referee with no such coupling.

\paragraph{Restricted-subset (non-degenerate) performance} To directly address whether the headline SegFormer-B0 result is inflated by trivially separable empty-mask cases, the failure-detection evaluation was repeated on the 1,046-image restricted subset in which both masks contain substantive predicted foreground (Section~\ref{sec:benchmark}). On this subset, Agreement Dice achieves an ROC-AUC of 0.876, compared with 0.960 on the full standardized benchmark. The 177 excluded images correspond overwhelmingly to genuine segmentation failures (ground-truth $DSC \approx 0$ for all excluded cases). Repeating this restricted evaluation for the Independent YOLO Referee on its own 975-image ``Both Non-Empty'' subset yields an ROC-AUC of 0.783, compared with 0.923 on the full standardized benchmark --- a proportionally larger reduction than observed for SegFormer-B0. This indicates that the same-architecture control referee's discriminative power is more strongly concentrated in the comparatively easy empty-mask cases than the cross-architecture SegFormer-B0 referee, consistent with the broader finding that cross-architecture diversity provides additional discriminative strength beyond model independence alone. We therefore report both full-benchmark and restricted figures explicitly for each referee: 0.960 and 0.923 as the headline results on the full standardized benchmark, and 0.876 and 0.783 as conservative, restricted estimates of RBQE's discriminative performance when the comparatively trivial empty-mask cases are excluded.

\paragraph{Descriptor sensitivity: an upper-bound comparison} The head-to-head comparison above deliberately fixes Agreement Dice across all four referees to ensure a fair, uniform comparison. To assess whether referee-specific descriptor selection would alter this ranking, Table~\ref{tab:bestdescriptor} reports the best-performing agreement descriptor selected retrospectively for each referee, with bootstrap validation (1,000 resamples). Because the strongest descriptor was chosen after observing performance, these best-of-five ROC-AUCs are not used as the primary head-to-head referee comparison (Table~\ref{tab:referees}); they instead illustrate the upper bound achievable by referee-specific descriptor selection. SegFormer-B0 and the Independent YOLO Referee are unaffected, since Agreement Dice is already their best descriptor; UNet++ improves modestly under Area Ratio (0.938 $\to$ 0.943), and MedSAM improves under Centroid Distance (0.863 $\to$ 0.873). In neither case does descriptor-specific selection change the overall ranking of the four referees established above.

\begin{table}[t]
\centering
\caption{Descriptor sensitivity analysis: best-performing agreement descriptor selected retrospectively for each referee, with bootstrap validation (1,000 resamples). Threshold-dependent metrics at the Youden-optimal operating threshold.}
\label{tab:bestdescriptor}
\footnotesize
\setlength{\tabcolsep}{3pt}
\resizebox{\textwidth}{!}{%
\begin{tabular}{llccccccc}
\toprule
Referee & Best descriptor & ROC-AUC & \begin{tabular}[c]{@{}c@{}}Bootstrap\\ mean\end{tabular} & 95\% CI & Acc. & Prec. & Rec. & F1 \\
\midrule
SegFormer-B0 & Agreement Dice & 0.960 & 0.960 & [0.947, 0.972] & 0.926 & 0.762 & 0.894 & 0.823 \\
UNet++ & Area Ratio & 0.943 & 0.942 & [0.925, 0.956] & 0.883 & 0.639 & 0.907 & 0.750 \\
Independent YOLO Referee & Agreement Dice & 0.923 & 0.923 & [0.902, 0.942] & 0.908 & 0.710 & 0.881 & 0.786 \\
MedSAM & Centroid Distance & 0.873 & 0.873 & [0.840, 0.908] & 0.952 & 1.000 & 0.750 & 0.857 \\
\bottomrule
\end{tabular}}
\end{table}

\subsection{RQ3: Robustness to the failure definition and to empty-mask cases}
\label{sec:rq3}

\begin{figure}[t]
\centering
\includegraphics[width=0.6\linewidth]{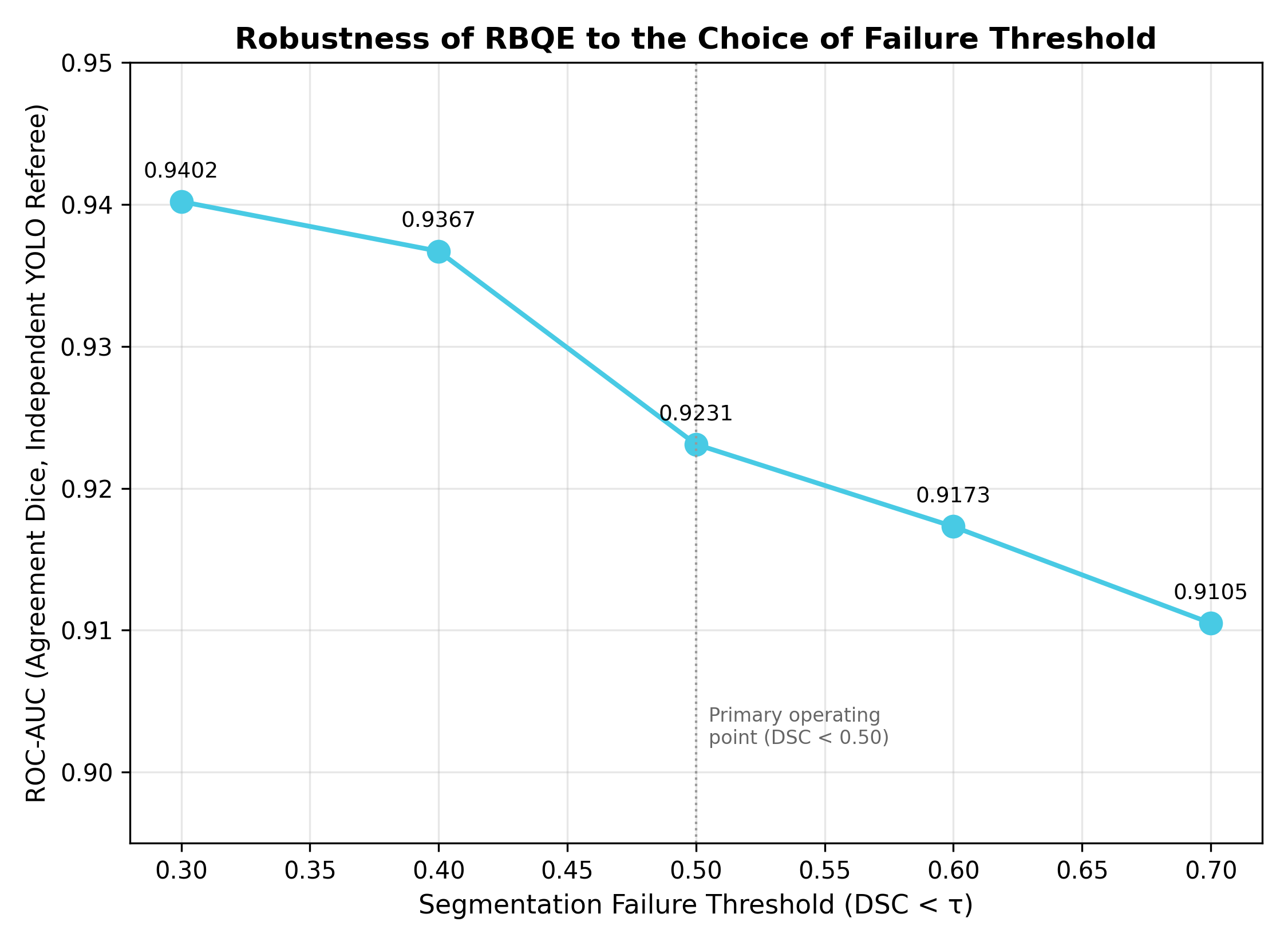}
\caption{Robustness of Agreement Dice failure-detection performance to the definition of segmentation failure. The failure threshold $\tau$ is varied from $DSC < 0.30$ to $DSC < 0.70$ using the Independent YOLO Referee, with $\tau = 0.50$ representing the primary operating point.}
\label{fig:threshold}
\end{figure}

\paragraph{Robustness to different failure definitions} The principal experiments define a segmentation failure as a prediction with $DSC < 0.50$. Although this operating point has been widely adopted in the segmentation quality estimation literature, the practical definition of a segmentation failure may vary depending on the clinical application: more conservative deployment scenarios may require higher segmentation quality, whereas other applications may tolerate lower segmentation accuracy. To investigate whether RBQE remains effective under different operating conditions, additional experiments were performed using multiple Dice thresholds for defining segmentation failure: 0.30, 0.40, 0.50, 0.60, and 0.70, with all other settings held fixed. The analysis used the Independent YOLO Referee, as the same-architecture control condition, and the Agreement Dice descriptor throughout, so that only the failure threshold was varied. Table~\ref{tab:thresholds} summarizes the results, and Fig.~\ref{fig:threshold} illustrates the variation in ROC-AUC across the evaluated failure definitions.

\begin{table}[t]
\centering
\caption{Performance of the proposed RBQE framework using the Independent YOLO Referee under different segmentation failure definitions, using the Agreement Dice descriptor.}
\label{tab:thresholds}
\small
\begin{tabular}{lccccc}
\toprule
Failure definition & ROC-AUC & Accuracy & Precision & Recall & F1 \\
\midrule
$DSC < 0.30$ & 0.940 & 0.926 & 0.729 & 0.910 & 0.809 \\
$DSC < 0.40$ & 0.937 & 0.911 & 0.693 & 0.914 & 0.788 \\
$DSC < 0.50$ & 0.923 & 0.908 & 0.710 & 0.881 & 0.786 \\
$DSC < 0.60$ & 0.917 & 0.910 & 0.747 & 0.859 & 0.799 \\
$DSC < 0.70$ & 0.911 & 0.901 & 0.782 & 0.801 & 0.791 \\
\bottomrule
\end{tabular}
\end{table}

Overall, the proposed framework exhibits stable performance across multiple failure thresholds, indicating that the agreement-based reliability signal is not restricted to a particular operating point. Lower Dice thresholds primarily identify severe segmentation failures; higher thresholds progressively include moderate-quality segmentations, making the classification problem harder as the distinction between successful and failed segmentations becomes less pronounced. Nevertheless, RBQE continues to provide meaningful discrimination across the evaluated range, confirming that its effectiveness does not depend on a single empirically selected Dice threshold and remains applicable under different clinical requirements.

\begin{figure}[t]
\centering
\includegraphics[width=0.6\linewidth]{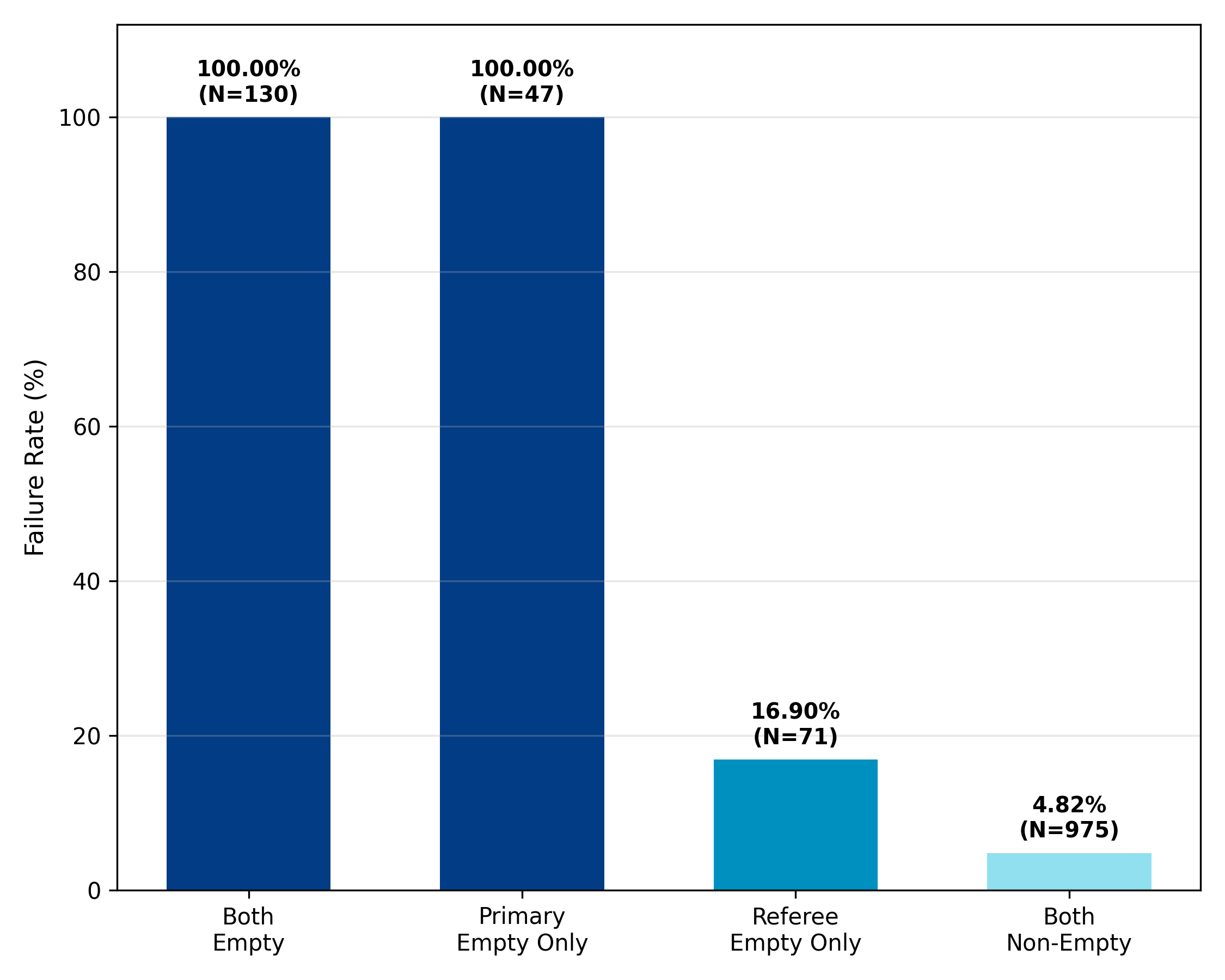}
\caption{Failure rate stratified by empty-mask status (Independent YOLO Referee, 1,223-image benchmark), corresponding to Table~\ref{tab:stratification}.}
\label{fig:stratified}
\end{figure}

\paragraph{Stratification by empty-mask status} To confirm that the framework's overall discriminative performance is not disproportionately driven by degenerate cases, the standardized 1,223-image benchmark, evaluated with the Independent YOLO Referee, was stratified into four mutually exclusive groups based on the emptiness of the primary and referee predictions (Table~\ref{tab:stratification}, Fig.~\ref{fig:stratified}). The four groups sum to exactly 1,223 images, confirming complete coverage of the benchmark. As expected, both ``Both Empty'' and ``Primary Empty Only'' cases correspond to complete segmentation failure (100\% failure rate), since an empty primary prediction cannot achieve meaningful overlap with a non-empty ground-truth lesion. The overwhelming majority of the benchmark (975 of 1,223 images, 79.7\%) falls into the ``Both Non-Empty'' group, where both models produced a non-trivial prediction; within this group, the true segmentation failure rate is only 4.82\%. Trivial empty-mask agreement is therefore not the dominant composition of the benchmark, and the restricted-subset results already reported in Section~\ref{sec:rq2} quantify RBQE's discriminative performance on precisely this non-degenerate regime.

\begin{table}[t]
\centering
\caption{Groupwise failure rate stratified by empty-mask status (Independent YOLO Referee, 1,223-image benchmark).}
\label{tab:stratification}
\small
\begin{tabular}{lrr}
\toprule
Group & Images & Failure rate \\
\midrule
Both Empty & 130 & 100.0\% \\
Primary Empty Only & 47 & 100.0\% \\
Referee Empty Only & 71 & 16.9\% \\
Both Non-Empty & 975 & 4.82\% \\
\bottomrule
\end{tabular}
\end{table}

\subsection{RQ4: Comparison with existing no-reference SQE methods}
\label{sec:rq4}

\begin{figure}[t]
\centering
\includegraphics[width=0.6\linewidth]{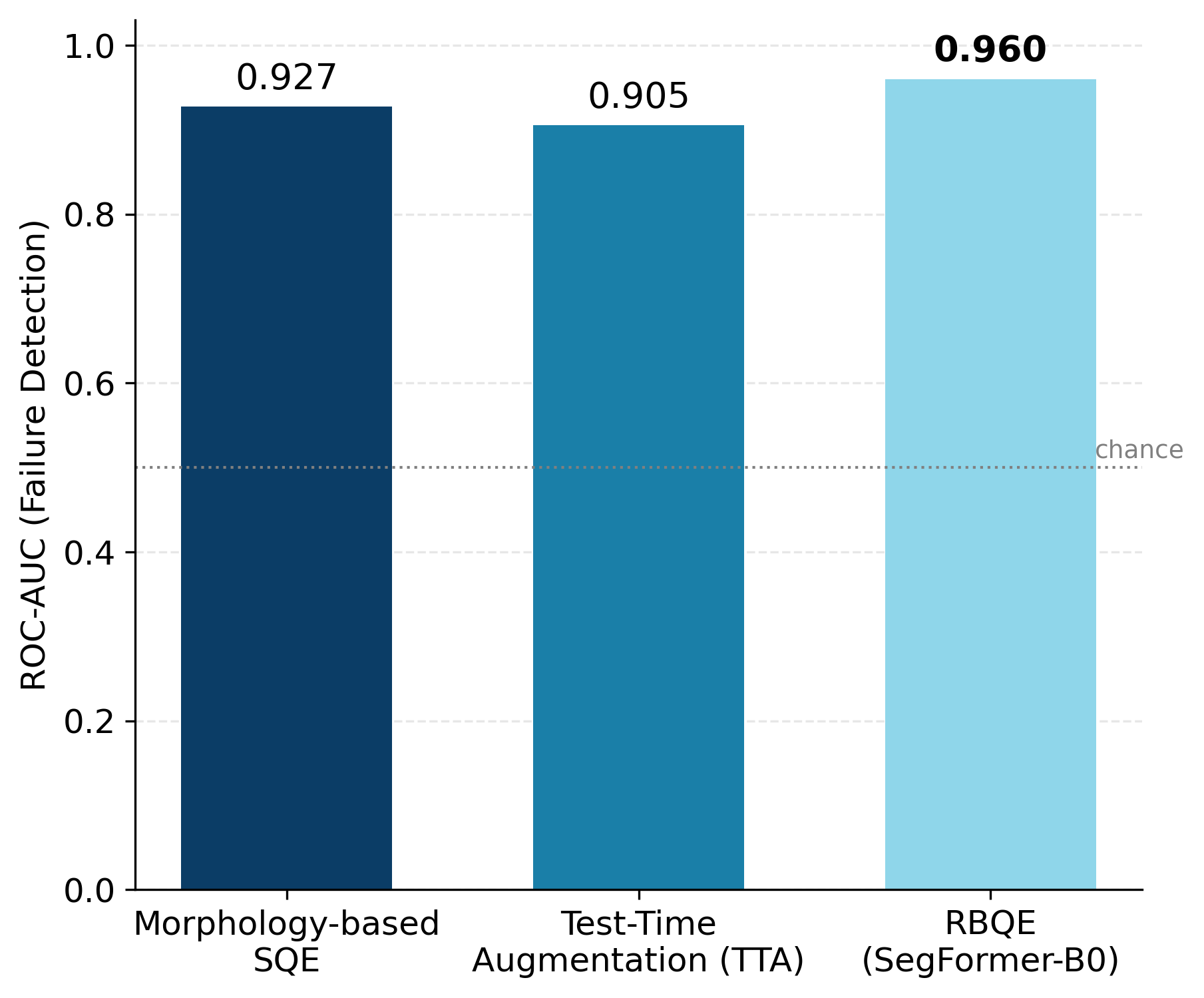}
\caption{Comparison of RBQE (SegFormer-B0 referee) against representative no-reference quality estimation paradigms --- morphology-based SQE and Test-Time Augmentation (TTA) --- evaluated under the identical protocol described in Section~\ref{sec:metrics}.}
\label{fig:baselines}
\end{figure}

Having established RBQE's effectiveness across referee configurations, we position the proposed framework against representative existing paradigms for no-reference segmentation quality estimation: morphology-based descriptors, which derive a quality signal from geometric properties of the predicted mask alone, and Test-Time Augmentation, which derives an uncertainty signal from a single model's stochastic behavior across repeated inference (Section~\ref{sec:baselines}). Both baselines were evaluated under the identical protocol, on the same standardized 1,223-image benchmark, and additionally on the identical 1,046-image restricted subset used for the SegFormer-B0 configuration. Table~\ref{tab:sqecomparison} and Fig.~\ref{fig:baselines} present the comparison.

\begin{table}[t]
\centering
\resizebox{\textwidth}{!}{%
\begin{threeparttable}
\caption{Comparison of RBQE (SegFormer-B0 referee) against representative no-reference quality estimation paradigms, evaluated on the full standardized 1,223-image benchmark and recomputed on the identical 1,046-image restricted (non-degenerate) subset.}
\label{tab:sqecomparison}
\small
\begin{tabular}{lcc}
\toprule
Method & ROC-AUC, full ($N = 1{,}223$) & ROC-AUC, restricted ($N = 1{,}046$) \\
\midrule
Morphology-based SQE & 0.927 & 0.709 \\
Test-Time Augmentation (TTA)\tnote{a} & 0.905 & 0.621 \\
RBQE (SegFormer-B0) & \textbf{0.960} & \textbf{0.876} \\
\bottomrule
\end{tabular}
\begin{tablenotes}\footnotesize
\item[a] TTA reaches ROC-AUC $=0.914$ on the pre-exclusion 1{,}248-image benchmark (provenance only, Section~\ref{sec:benchmark}); the value shown here (0.905) uses the standardized 1{,}223-image benchmark, matching the protocol applied to RBQE and the morphology baseline throughout.
\end{tablenotes}
\end{threeparttable}}
\end{table}

On the full standardized benchmark, RBQE with the SegFormer-B0 referee achieved a ROC-AUC of 0.960, compared with 0.905 for the TTA baseline and 0.927 for the best-performing morphology-based baseline (Circularity), exceeding the protocol-matched TTA baseline by 0.055 ROC-AUC and the morphology-based baseline by 0.033 ROC-AUC. These results are consistent with the central hypothesis of this work: single-model uncertainty signals such as TTA remain bound to the calibration and stochastic behavior of the model being interrogated, whereas RBQE derives its reliability signal from agreement between two independently trained segmentation models. The morphology-based baseline demonstrates that geometric properties of the predicted mask contain useful information about segmentation quality; however, such descriptors cannot identify failures in which a predicted mask remains geometrically plausible while being incorrectly localized or delineated relative to the true lesion.

The restricted-subset comparison is more discriminating still. Both baselines degrade substantially more than RBQE when trivially separable empty-mask cases are excluded: morphology-based SQE falls from 0.927 to 0.709, and TTA falls from 0.905 to 0.621, whereas RBQE (SegFormer-B0) falls from 0.960 to 0.876. On the identical, protocol-matched restricted subset, RBQE therefore exceeds the morphology baseline by 0.167 ROC-AUC and TTA by 0.255 ROC-AUC --- a substantially wider margin than on the full benchmark (0.033 and 0.055, respectively). This pattern is consistent with both baselines' comparative reliance on the same easily detected empty-mask cases that RBQE's restricted evaluation deliberately excludes: circularity is undefined or trivially informative for an empty mask, and TTA's zero-prediction-count signal is, by construction, maximally informative precisely when the primary model produces no prediction at all. RBQE's agreement-based signal, by contrast, retains substantial discriminative power even when restricted to cases where the primary model does produce a non-trivial prediction. (The Independent YOLO Referee's restricted result, 0.783 on its own 975-image subset, was not benchmarked against the restricted baselines, since its restricted subset differs in composition from SegFormer-B0's; the comparison in Table~\ref{tab:sqecomparison} is specific to the headline SegFormer-B0 configuration.)

These results indicate that agreement-based reliability estimation is not merely competitive with, but measurably stronger than, both a purely geometric baseline and a representative single-model uncertainty method, while requiring only one additional referee forward pass at deployment.

\subsection{Deployment characteristics: selective prediction and computational cost}
\label{sec:deployment}

\begin{figure}[t]
\centering
\includegraphics[width=0.8\linewidth]{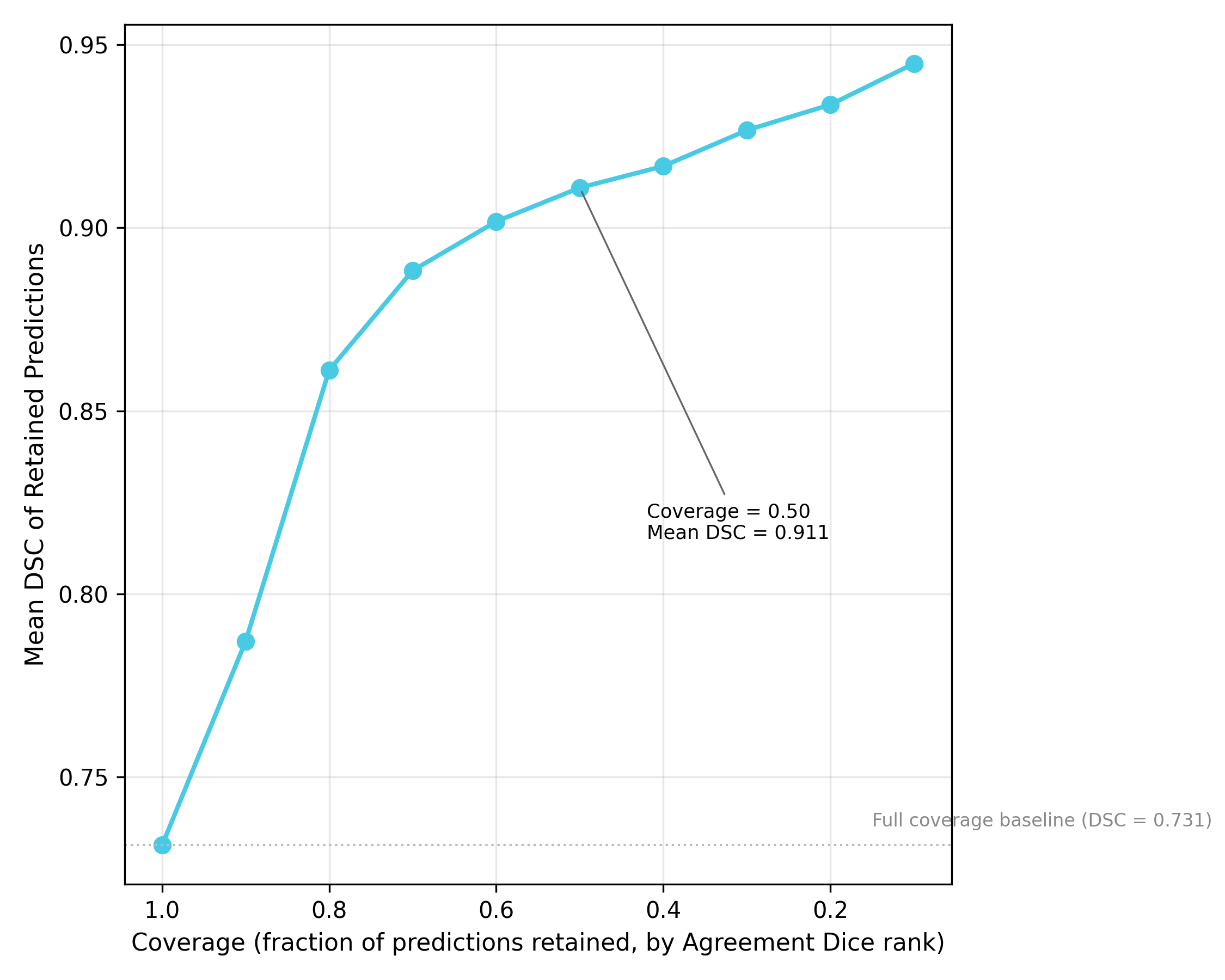}
\caption{Risk--coverage curve for selective prediction using Agreement Dice (Independent YOLO Referee, 1,223-image benchmark). Rejecting lower-agreement predictions monotonically increases the mean DSC of the retained predictions, from 0.731 at full coverage to 0.945 at 10\% coverage.}
\label{fig:riskcov}
\end{figure}

\paragraph{Risk--coverage analysis} The preceding sections evaluate RBQE as a binary failure detector at a fixed operating threshold. In practice, RBQE's continuous agreement score also supports selective prediction: rather than accepting every prediction, a deployment system can reject the lowest-agreement predictions for expert review, retaining only those above a chosen confidence level. We evaluate this use case via a risk--coverage analysis, performed using the Independent YOLO Referee to assess whether the continuous agreement signal supports selective prediction under the same-architecture control condition: predictions are ranked by Agreement Dice and progressively rejected from lowest to highest agreement, reporting the mean DSC of the retained predictions at each coverage level. Fig.~\ref{fig:riskcov} and Table~\ref{tab:riskcov} present the complete risk--coverage curve.

\begin{table}[t]
\centering
\caption{Risk--coverage analysis: mean DSC of retained predictions as coverage decreases, ranked by Agreement Dice (Independent YOLO Referee, 1,223-image benchmark).}
\label{tab:riskcov}
\small
\begin{tabular}{cc}
\toprule
Coverage & Mean DSC \\
\midrule
1.00 & 0.731 \\
0.90 & 0.787 \\
0.80 & 0.861 \\
0.70 & 0.888 \\
0.60 & 0.902 \\
0.50 & 0.911 \\
0.40 & 0.917 \\
0.30 & 0.927 \\
0.20 & 0.934 \\
0.10 & 0.945 \\
\bottomrule
\end{tabular}
\end{table}

Mean DSC increases monotonically as coverage decreases, from 0.731 at full coverage to 0.945 when only the top 10\% of predictions by Agreement Dice are retained. The full-coverage mean DSC of 0.731 is higher than the overall mean DSC of 0.717 reported for the 1,248-image external benchmark in Table~\ref{tab:primaryperf} because the risk--coverage analysis uses the standardized 1,223-image RBQE benchmark after exclusion of the 25 images removed during benchmark construction (Section~\ref{sec:benchmark}). At a coverage of 0.50 --- rejecting the least reliable half of predictions for review --- the mean DSC of the retained predictions reaches 0.911, representing a substantial improvement over the full-coverage value. This confirms that Agreement Dice is not only an effective binary failure detector, but also a useful continuous ranking signal: predictions it deems more reliable are, on average, genuinely of higher quality, supporting its use as a practical triage mechanism in clinical deployment where a proportion of predictions can be flagged for expert review rather than accepted automatically.

\paragraph{Computational complexity and deployment feasibility} Practical deployment of segmentation quality estimation methods also requires consideration of computational efficiency and inference cost. Unlike uncertainty estimation methods based on TTA, which require multiple forward passes through the same segmentation model, RBQE performs only a single forward pass using the primary model and one additional forward pass using an independently trained referee model. The agreement descriptors are subsequently computed using simple, deterministic geometric operations on binary segmentation masks, introducing negligible computational cost relative to the segmentation inference itself and requiring no optimization or additional learnable parameters during deployment.

\begin{figure}[t]
\centering
\includegraphics[width=0.80\linewidth]{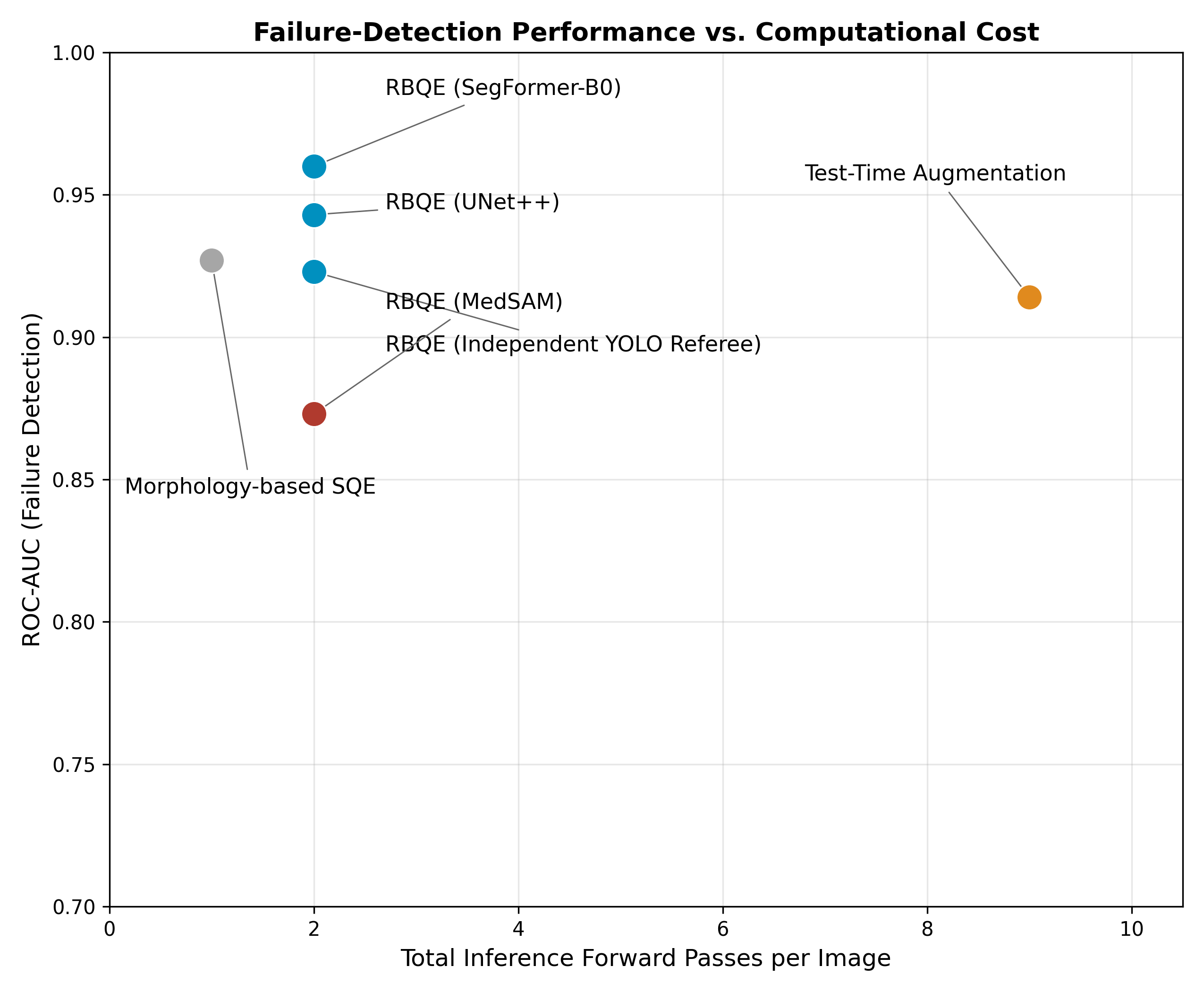}
\caption{Relationship between segmentation failure-detection performance (ROC-AUC) and the total number of inference forward passes required per image by the evaluated deployment-time quality estimation methods. Morphology-based quality estimation requires one primary-model inference, TTA requires nine augmented primary-model inferences, and RBQE requires one primary-model inference plus one referee-model inference.}
\label{fig:cost}
\end{figure}

Table~\ref{tab:computational} summarizes the computational and practical deployment characteristics of the evaluated quality estimation approaches, including all four RBQE referee configurations; Fig.~\ref{fig:cost} illustrates the relationship between failure-detection performance and inference forward-pass burden. Overhead is reported qualitatively, in terms of the number and nature of forward passes required, rather than as hardware-specific latency, since absolute inference time depends on GPU, batch size, precision, and implementation details that were not held constant across methods in this study.

\begin{table}[t]
\centering
\caption{Computational and deployment comparison of no-reference segmentation quality estimation methods.}
\label{tab:computational}
\footnotesize
\resizebox{\textwidth}{!}{%
\begin{tabular}{lcccc}
\toprule
Method & Additional model & Forward passes & Quality labels required & Deployment overhead \\
\midrule
Morphology-based SQE & No & 1 & No & Very low \\
Test-Time Augmentation & No & Multiple & No & High \\
RBQE (Independent YOLO Referee) & Yes & 2 & No & Moderate \\
RBQE (SegFormer-B0) & Yes & 2 & No & Moderate \\
RBQE (UNet++) & Yes & 2 & No & Moderate \\
RBQE (MedSAM) & Yes & 2 & No & High \\
\bottomrule
\end{tabular}}
\end{table}

Among the evaluated referee models, the Independent YOLO Referee, SegFormer-B0, and UNet++ require one additional forward pass using lightweight segmentation architectures. In contrast, MedSAM operates as a frozen foundation model at a substantially higher input resolution ($1024 \times 1024$), resulting in greater inference cost despite eliminating the need for referee training. The choice of referee architecture therefore provides flexibility in balancing accuracy and computational efficiency according to application requirements: lightweight, same-architecture referees may be preferred for real-time or resource-constrained systems; cross-architecture referees such as SegFormer-B0 offer the strongest failure-detection performance at moderate additional cost; and foundation-model referees such as MedSAM may be suited to offline analysis where training data or infrastructure for a dedicated referee is unavailable, despite their higher inference latency. A further practical advantage of RBQE is its modular design: since the referee model operates independently of the primary segmentation network, existing segmentation systems can be augmented with reliability estimation without modifying the original segmentation architecture or retraining the deployed model, simplifying integration into clinical workflows and facilitating compatibility with heterogeneous segmentation models developed by different research groups or manufacturers.

\subsection{Qualitative analysis}
\label{sec:qualitative}

\begin{figure}[t]
\centering
\includegraphics[width=0.95\linewidth]{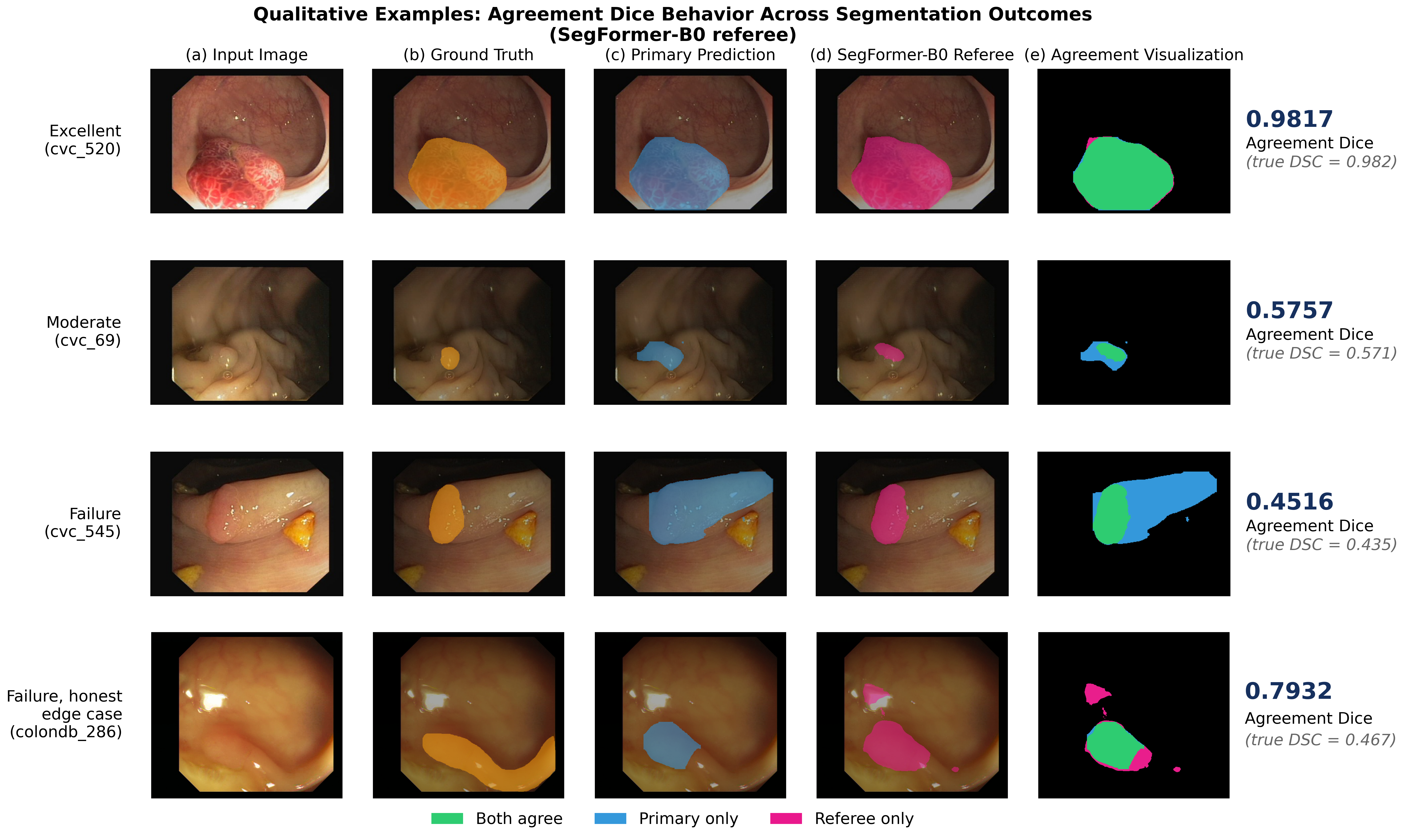}
\caption{Representative qualitative examples illustrating the behavior of Agreement Dice across successful (image cvc\_520), moderately challenging (image cvc\_69), and failed (image cvc\_545) segmentation cases, together with one additional high-agreement failure mode (bottom, image colondb\_286) in which the primary and referee models agree closely (Agreement Dice $= 0.793$) while both diverge substantially from the ground truth ($DSC = 0.467$), illustrating a genuine limitation of agreement-based reliability estimation. Cases are drawn from CVC-ClinicDB (rows 1--3) and CVC-ColonDB (row 4). (a) Input image; (b) ground truth; (c) primary model prediction; (d) SegFormer-B0 referee prediction; (e) agreement visualization (green $=$ both models agree, blue $=$ primary only, magenta $=$ referee only).}
\label{fig:qualitative}
\end{figure}

While quantitative evaluation demonstrates the overall effectiveness of the proposed framework, qualitative analysis provides additional insight into how cross-model agreement relates to segmentation reliability across different clinical scenarios. Fig.~\ref{fig:qualitative} presents four representative examples, one per row, spanning the range of segmentation outcomes observed in the external benchmark. For each example: (a) the input colonoscopy image, (b) the ground-truth annotation, (c) the primary model's predicted segmentation, (d) the SegFormer-B0 referee's predicted segmentation, (e) a visualization of the agreement between the two masks, and (f) the resulting Agreement Dice value are shown.

In the high-quality segmentation case (Fig.~\ref{fig:qualitative}, first row), the primary and referee models produce highly consistent segmentation masks with substantial overlap and similar boundary localization; consequently, the computed Agreement Dice value is high, correctly reflecting the successful segmentation outcome. The moderately challenging case (second row) exhibits partial disagreement between the two models: although the overall lesion location remains consistent, differences arise along the object boundary and in the predicted lesion extent, producing an intermediate Agreement Dice value that reflects the increased uncertainty associated with this prediction. In contrast, the failed segmentation case (third row) is characterized by pronounced disagreement between the independently trained models, manifesting as incomplete lesion delineation, inaccurate boundary localization, or a missed polyp; as a result, the Agreement Dice value decreases substantially, enabling RBQE to correctly identify this segmentation as unreliable.

An important observation across these examples is that disagreement does not necessarily arise from identical error patterns. In several challenging cases, one model produces a reasonable segmentation while the other fails, leading to low agreement that serves as an effective indicator of reduced prediction reliability. This behavior demonstrates that RBQE captures complementary information from independent segmentation models rather than relying solely on the confidence of a single predictor.

The fourth row presents a case that RBQE does not handle well, included deliberately for balance. Here the primary and referee models agree closely with each other (Agreement Dice $= 0.793$) despite both diverging substantially from the ground truth ($DSC = 0.467$): the two models converge on a similar, but similarly wrong, delineation of the lesion. This is precisely the failure mode anticipated by the framework's reliance on independent failure modes rather than independent training alone --- when two models share a systematic bias toward a particular imaging condition or lesion appearance, their agreement can no longer be assumed to track correctness. We return to this limitation in Section~\ref{sec:limitations}.

\section{Discussion}
\label{sec:discussion}

The experiments presented in this study demonstrate that agreement between independently trained segmentation models provides an effective, low-cost surrogate for segmentation reliability at deployment. Across the agreement descriptors, four referee configurations, and a standardized external benchmark of 1,223 colonoscopy images --- with all headline results additionally verified on non-degenerate restricted subsets --- cross-model agreement consistently distinguished reliable from unreliable segmentations without requiring access to ground-truth annotations at inference.

\paragraph{Why does cross-model agreement work?} The strongest agreement descriptors in this study --- Agreement Dice and Agreement IoU (ROC-AUC $= 0.960$), followed by Centroid Distance (0.951) and Boundary Agreement (0.949) --- are directly derived from spatial and geometric consistency between two separately parameterized segmentation models. Although the models are trained using the same development dataset, they do not share weights, checkpoints, or optimization state, and cross-architecture referees additionally introduce different architectural inductive biases. Consequently, agreement between their predictions can provide an additional check on the primary model's reliability, although shared dataset exposure and systematic biases can still produce correlated errors --- as illustrated by the high-agreement failure case in Section~\ref{sec:qualitative}. The results of Section~\ref{sec:rq2} give this mechanism a precise formulation: model independence is the necessary foundation of the agreement signal, architectural diversity a further --- though not strictly required --- source of additional discriminative strength, and output-level coupling (as in MedSAM's prompt dependence) is sufficient to undermine the signal even under maximal architectural diversity.

\paragraph{Positioning relative to existing no-reference approaches} Section~\ref{sec:rq4} substantiates, empirically, the taxonomy introduced in Sections~\ref{sec:intro}--\ref{sec:related}: morphology-based descriptors capture only geometric regularity and cannot detect plausible-looking but incorrectly localized predictions, while TTA, like other single-model uncertainty methods, remains fundamentally bound to the calibration of one model and is therefore vulnerable precisely when that model is confidently wrong --- a vulnerability made explicit by the sharp degradation of both baselines on the non-degenerate restricted subset. Unlike RCA \citep{Valindria2017}, which requires fitting a new reverse classifier for every test image against a reference atlas, RBQE requires no per-image retraining and no atlas dependency; unlike deep ensembles, which derive disagreement from same-architecture models differing only in initialization, RBQE's strongest results arise specifically from cross-architecture referees, indicating that architectural diversity provides genuine additional signal beyond what ensemble-style disagreement alone would offer.

\paragraph{Clinical relevance} During real-time colonoscopy, ground-truth segmentation is, by definition, unavailable, and learned segmentation accuracy cannot be directly measured at the point of use. RBQE addresses this gap by providing a reference-free, post-hoc reliability signal that can be computed alongside the primary model's prediction using one additional referee forward pass, without requiring quality-labelled training data. Beyond binary failure detection, the risk--coverage analysis (Section~\ref{sec:deployment}) shows that rejecting the lowest-agreement half of predictions raises mean retained DSC from 0.731 to 0.911, and retaining only the top 10\% raises it further to 0.945: RBQE can therefore operate not only as a binary alarm, but as a triage mechanism, flagging a tunable proportion of the least reliable predictions for expert review while allowing high-agreement predictions to proceed with greater confidence, offering clinicians and system designers explicit control over the trade-off between automation coverage and diagnostic risk. This positions RBQE not as a replacement for the primary segmentation model, but as an independent safety layer capable of flagging potentially unreliable predictions before they inform diagnostic or therapeutic decisions --- a role that is particularly valuable given that segmentation failures are most likely to occur under precisely the challenging imaging conditions (poor illumination, small polyps, domain shift) in which clinicians can least afford an undetected error.

\section{Limitations}
\label{sec:limitations}

Although the proposed RBQE framework demonstrates strong and consistent performance across referee configurations, failure thresholds, and robustness analyses, several limitations should be acknowledged.

First, this study evaluates RBQE exclusively on binary polyp segmentation. While this represents an important and clinically relevant task, the generalizability of agreement-based reliability estimation to multiclass segmentation problems, or to other medical imaging modalities beyond colonoscopy, remains to be established.

Second, RBQE's effectiveness depends on the availability of a suitable referee model. Although this study demonstrates that both same-architecture independence and cross-architecture diversity contribute to a useful signal (Section~\ref{sec:rq2}), deployment in a new clinical setting requires either training an additional referee model or identifying a suitable pretrained one, which represents a practical prerequisite not required by single-model uncertainty methods. Relatedly, the framework's core assumption --- that independently trained models fail in different ways --- can itself fail, as demonstrated by the high-agreement failure case in Fig.~\ref{fig:qualitative} (fourth row). Such cases are, by construction, difficult for any agreement-based method to detect, since the signal RBQE relies on is precisely the one that fails to distinguish them.

Third, the agreement descriptors employed in this study are deterministic geometric measures computed directly from binary masks. While effective, this design does not incorporate probabilistic segmentation outputs, feature-space similarity, or topology-aware representations, any of which may further improve the sensitivity of the framework to subtler failure modes.

Fourth, this evaluation, although comprising 1,223 external images across four independent public benchmarks, is retrospective. Prospective validation on multi-center clinical data, with variability in acquisition hardware and patient populations beyond that captured by the public benchmarks used here, is necessary before routine clinical deployment.

Finally, RBQE estimates the reliability of a segmentation prediction as a whole, but does not localize which region of a prediction is unreliable. Combining agreement-based reliability estimation with localized, pixel-level uncertainty maps may provide more actionable feedback for clinicians and downstream decision-support systems.

\section{Conclusion}
\label{sec:conclusion}

This work introduced Referee-Based Quality Estimation (RBQE), a deployment-time, reference-free framework that estimates segmentation reliability through agreement between a primary segmentation model and an independently trained referee, requiring no ground-truth annotations, no quality-labelled training data, and no modification of the deployed segmentation model.

A central finding of this study is the explicit separation of model independence from architectural diversity as contributors to the agreement signal. The Independent YOLO Referee --- identical in architecture to the primary model, differing only in initialization --- achieved a ROC-AUC of 0.923, confirming that independence alone is sufficient for a meaningful reliability signal. Cross-architecture referees provided a further, statistically significant improvement, with SegFormer-B0 achieving the strongest overall performance (0.960). This distinction, together with the identification of MedSAM's prompt-dependence as a case where architectural diversity does not translate into genuine output-level independence, provides a more precise account of why agreement-based quality estimation works than has previously been established in the literature.

Across a standardized external benchmark of 1,223 colonoscopy images spanning four independent public datasets, RBQE's performance was shown to be robust to the choice of failure threshold and to the choice of referee architecture. Discriminative performance declines when trivial empty-mask agreement cases are excluded (ROC-AUC falls from 0.960 to 0.876 for SegFormer-B0, and from 0.923 to 0.783 for the Independent YOLO Referee, when each is restricted to images with substantive foreground in both the primary and referee masks), confirming that part of the full-benchmark result is attributable to these comparatively easy cases; on the identical restricted subset, however, RBQE's margin over the evaluated baselines widens rather than narrows. RBQE was further shown to outperform representative morphology-based and TTA-based baselines by 0.033 and 0.055 ROC-AUC respectively under an identical evaluation protocol (Section~\ref{sec:rq4}), to support effective selective prediction via risk--coverage analysis, and to require substantially lower computational overhead than TTA-based alternatives, needing only one additional deterministic forward pass at inference.

Taken together, these results support agreement between independently trained segmentation models as a practical, interpretable, and computationally efficient reliability signal for deployment-time segmentation quality estimation, offering a potentially practical path toward integrating automated polyp segmentation reliability estimation into future colonoscopy workflows. Future work should extend this framework to multiclass and multi-modality segmentation tasks, incorporate richer agreement representations, and pursue prospective multi-center clinical validation.

\section*{Code availability}
The implementation of the proposed Referee-Based Quality Estimation (RBQE) framework, including the reproducibility notebook, evaluation scripts, and sample experimental results, is available at: \url{https://github.com/sidgupta307/RBQE-Referee-Based-Quality-Estimation}.

\section*{Data availability}
The external evaluation datasets used in this study (CVC-ClinicDB, CVC-ColonDB, ETIS-Larib PolypDB, CVC-300, and Kvasir-SEG) are publicly available third-party datasets, cited in Section~\ref{sec:datasets}. The code, trained model configurations, and derived experimental results supporting this study are available at the repository referenced in the Code availability section.

\section*{Ethics statement}
This study did not involve new data collection from human participants. All colonoscopy image datasets used (Kvasir-SEG, CVC-ClinicDB, CVC-ColonDB, ETIS-Larib PolypDB, CVC-300) are publicly available, pre-existing, de-identified datasets released by their original authors for research use; no new ethics approval was required for this work.

\section*{Declaration of competing interest}
The authors declare that they have no known competing financial interests or personal relationships that could have appeared to influence the work reported in this paper.

\section*{Funding}
This work was supported by the IITR@175 Fellowship, Indian Institute of Technology Roorkee.

\section*{Declaration of generative AI and AI-assisted technologies in the writing process}
During the preparation of this work, the author(s) used Claude (Anthropic) in order to improve the grammar, language, and readability of the manuscript text. After using this tool, the author(s) reviewed and edited the content as needed and take full responsibility for the content of the publication. All experimental results, analysis, and conclusions are the authors' own.

\section*{CRediT authorship contribution statement}
\textbf{Siddharth Gupta:} Conceptualization, Methodology, Software, Formal analysis, Investigation, Writing -- original draft. \textbf{Jitin Singla:} Conceptualization, Supervision, Writing -- review and editing.


\bibliographystyle{elsarticle-num}
\bibliography{refs}

\end{document}